\documentclass[nohyperref]{article}

\usepackage{microtype}
\usepackage{graphicx}
\usepackage{subfigure}
\usepackage{booktabs} 

\usepackage{hyperref}

\usepackage[accepted]{icml2022}

\usepackage{amsmath}
\usepackage{amssymb}
\usepackage{mathtools}
\usepackage{amsthm}

\usepackage{amsmath,amsfonts,bm}

\def\eqref#1{equation~\ref{#1}}

\def\1{\bm{1}}

\DeclareMathAlphabet{\mathsfit}{\encodingdefault}{\sfdefault}{m}{sl}
\SetMathAlphabet{\mathsfit}{bold}{\encodingdefault}{\sfdefault}{bx}{n}

\newcommand{\E}{\mathbb{E}}

\newcommand{\R}{\mathbb{R}}

\usepackage{url}

\usepackage{booktabs}       
\usepackage{amsfonts}       
\usepackage{nicefrac}       
\usepackage{microtype}      

\usepackage{multirow}
\usepackage{graphicx}
\usepackage{subfigure}
\usepackage{wrapfig}

\usepackage[capitalize,noabbrev]{cleveref}

\theoremstyle{plain}
\newtheorem{theorem}{Theorem}

\theoremstyle{definition}
\newtheorem{definition}{Definition}
\newtheorem{assumption}{Assumption}
\theoremstyle{remark}

\global\long\def\th{\theta}%
\global\long\def\w{\boldsymbol{w}}%
\global\long\def\L{\mathcal{L}}%
\global\long\def\E{\mathbb{E}}%
\global\long\def\P{\mathbb{P}}%
\global\long\def\N{\mathcal{N}}%
\global\long\def\W{\mathcal{W}}%
\global\long\def\LL{\textbf{L}}%

\usepackage[textsize=tiny]{todonotes}

\icmltitlerunning{Centroid Approximation for Bootstrap}

\begin{document}
\twocolumn[
\icmltitle{Centroid Approximation for Bootstrap: Improving Particle Quality at Inference}



\icmlsetsymbol{equal}{*}

\begin{icmlauthorlist}
\icmlauthor{Mao Ye}{yyy}
\icmlauthor{Qiang Liu}{yyy}
\end{icmlauthorlist}

\icmlaffiliation{yyy}{Department of Computer Science, University of Texas at Austin}

\icmlcorrespondingauthor{Mao Ye}{maoye21@utexas.edu}

\icmlkeywords{Machine Learning, ICML}

\vskip 0.3in
]



\printAffiliationsAndNotice{}  

\begin{abstract}
Bootstrap is a principled and powerful frequentist statistical tool for uncertainty quantification. Unfortunately, standard bootstrap methods are computationally intensive due to the need of drawing a large  i.i.d. bootstrap sample to approximate the ideal bootstrap distribution; this largely hinders their application in large-scale machine learning, especially deep learning problems. In this work, we propose an efficient method to explicitly \emph{optimize} a small set of high quality  ``centroid'' points to better approximate 
the ideal bootstrap distribution. We achieve this by minimizing a simple objective function that is asymptotically equivalent to the Wasserstein distance to the ideal bootstrap distribution. This allows us to provide an accurate estimation of uncertainty with a small number of bootstrap centroids,  outperforming the  naive i.i.d. sampling approach. 
Empirically, we show that our method can boost the performance of bootstrap in a variety of applications.
\end{abstract}

\section{Introduction}
Bootstrap is a simple and principled frequentist uncertainty quantification tool and can be flexibly applied to obtain data uncertainty estimation with strong theoretical guarantees  \citep{hall1988rate,austern2020asymptotics,chatterjee2005generalized,cheng2010bootstrap}.  
In particular, when combined with the maximum likelihood estimator or more general M-estimators, bootstrap provides a general-purpose, plug-and-play non-parametric inference framework for general probabilistic models without case-by-case derivations; this makes it a promising frequentist alternative to Bayesian inference. 

However, the standard bootstrap inference is highly expensive in both computation and memory as it typically requires drawing a large number\footnote{For example, thousands of, as suggested by Statistics textbooks such as \citet{wasserman2013all}.} of i.i.d. bootstrap particles (samples) to obtain an accurate uncertainty estimation. In the context of this paper, as each bootstrap particle/sample/centroid is a machine learning model, we might directly call a model as particle/sample/centroid. With a small number of particles, bootstrap may perform poorly. As a consequence, when applied to deep learning, we need to store a large number of neural networks and feed the input into a tremendous number of networks every time we make inference, which can be quite expensive and even unaffordable for deep learning problems with huge models. While training cost is an extra burden, it is small compared with the cost of making prediction as we only need to train the model once but make countless predictions at deployment. For example, in autonomous driving applications, our device can only store a limited number of models and we need to make decisions within a short time, which makes the standard bootstrap with a large number of models no more feasible. Typical ensemble methods in deep learning, such as  \citet{lakshminarayanan2016simple,huang2017snapshot,vyas2018out, maddox2019simple, liu2016stein}, can only afford to use a small number (e.g., less than 20) of models.

Therefore, to make bootstrap more accessible in modern machine learning, it is essential to develop new approaches that break the key computation and memory barriers mentioned above. We are motivated to consider the following problem:

\emph{How to improve the accuracy of bootstrap when the number of particles \textbf{at inference} is limited?}

Here we emphasis that our main goal is not reducing the training time but improve the particle quality for inference. We attack this challenge by presenting an efficient centroid approximation for bootstrap. Our method replaces the i.i.d. bootstrap particles with a set of carefully optimized \emph{centroid particles} that are guaranteed to provide an accurate and compact approximation to the ideal bootstrap distribution so that only a smaller number of particles is needed to obtain good performance.

\begin{figure}
\begin{center}
\includegraphics[scale=0.0665]{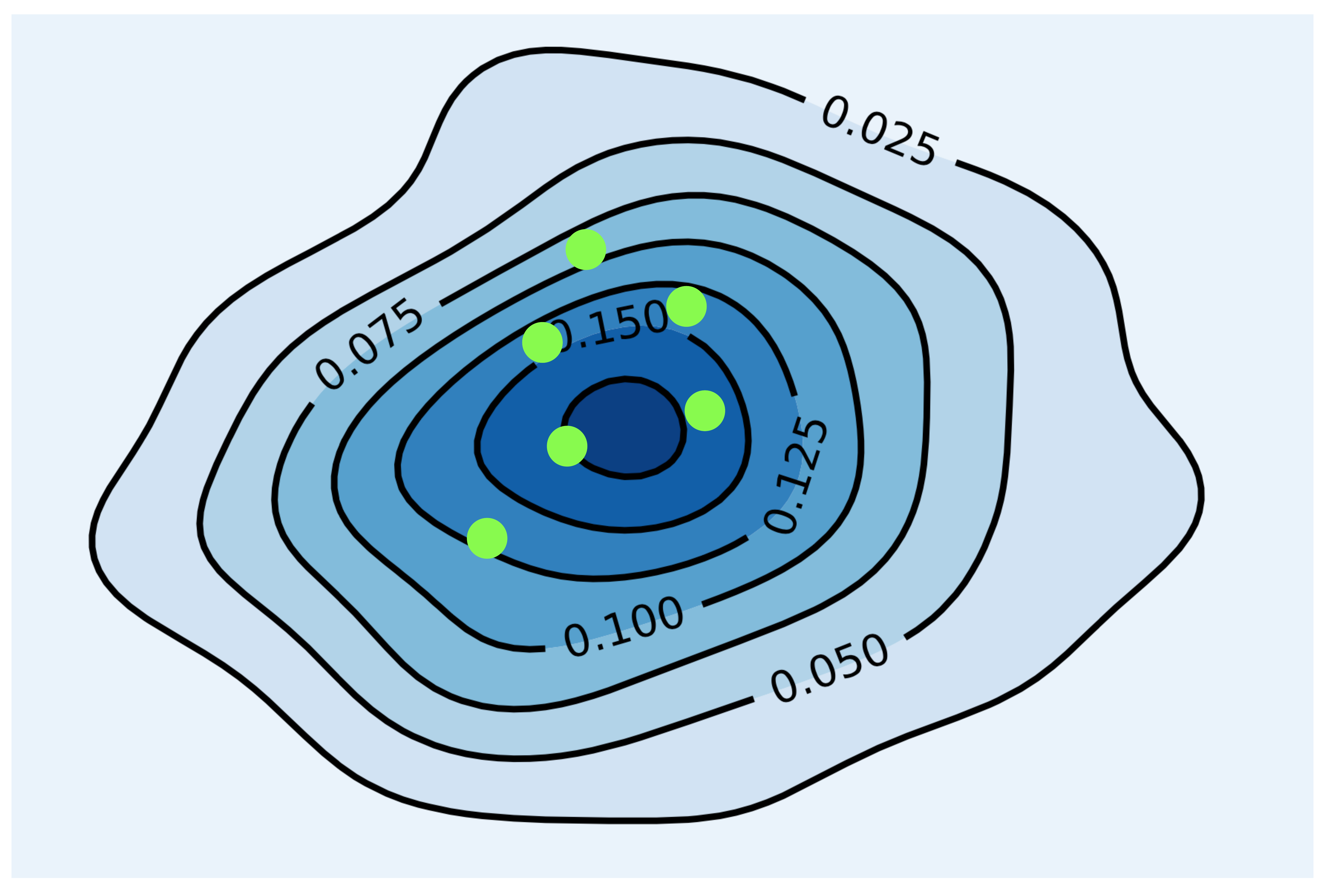}
\includegraphics[scale=0.0665]{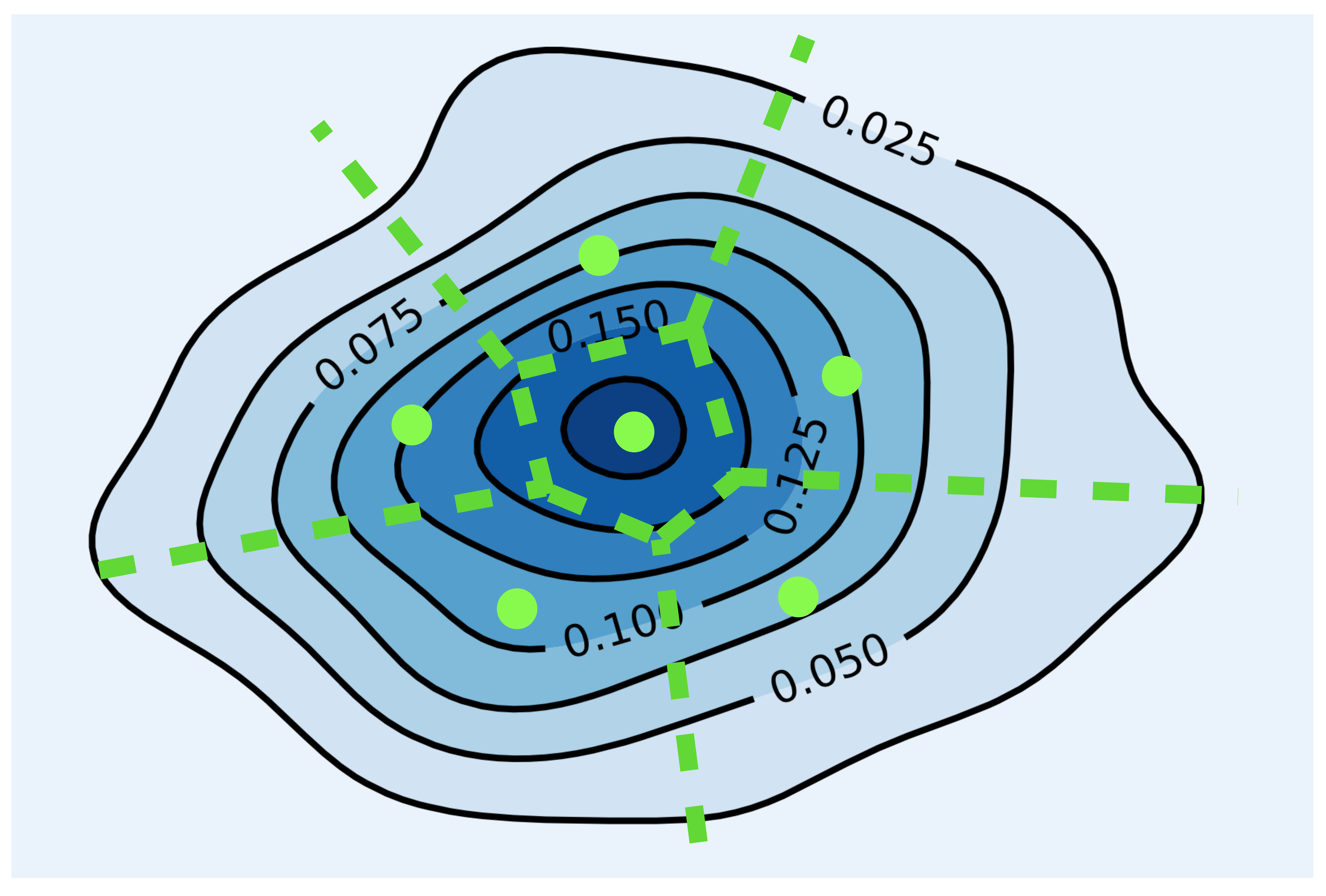}
\caption{The solid lines represent the density of the target distribution. Left figure: Typical i.i.d. particles that are randomly distributed on the domain. Right figure: The learned diversified centroids that are well distributed on the domain. The centroids partition the domain into several disjoint regions (separated by the dashed lines in the figure) and each centroid can be viewed as the `K-means' center of the region it belongs to.} \label{fig: illustration}
\end{center}
\end{figure}

Our method is based on minimizing a specially designed objective function that 
is asymptotically equivalent to the Wasserstein distance between the ideal bootstrap distribution and the particle distribution formed by the learned centroids. During the training, each centroid adjusts its location being aware of the locations of the others so that centroids are diversified and well distributed on the domain. Our method is similar to doing K-means on the ideal bootstrap distribution, finding K representative centroids that well represent K separate parts of the target distribution’s domain in an optimal way. As centroids are optimized to better approximate the distribution, our approach naturally improves over the vanilla bootstrap with i.i.d. particles. See Figure \ref{fig: illustration} for illustration.

Empirically, we apply the centroid approximation method to various applications, including confidence interval estimation \citep{diciccio1996bootstrap}, bootstrap method for contextual bandit \citep{riquelme2018deep}, bootstrap deep Q-network \citep{osband2016deep} and bagging method \citep{breiman1996bagging} for neural networks. We find that our method consistently improves over the standard bootstrap.

\paragraph{Notation}
We use $\left\Vert \cdot\right\Vert$ to represent the $\ell_{2}$ norm
for a vector and the operator norm for a matrix. We denote the integer set $\{1,2,....,N\}$ by $[N]$. Given any $m$, we define the probability simplex $\mathcal{C}^{m}:=\{[v_{1},...,v_{m}]\in\mathbb{R}^{m}:v_{i}\ge0,\ \forall i\in[m]\ \text{and}\ \sum_{i\in[m]}v_{i}=1\}$.
For a symmetric matrix $M$, we denote its minimal eigenvalue by $\lambda_{\min}(M)$. 
For a positive-definite matrix $M$, if $M=A^\top A$, then we denote $A$ by $M^{1/2}$. We denote the Wasserstein distance between two distribution $\rho_1$ and $\rho_2$ by $\W_{2}[\rho_1,\rho_2]$.
We use $O$ and $o$ to denote the conventional big-O and small-o notation and use $O_{p}$ to denote the stochastic boundedness. We use $\overset{d}{\to}$ to denote convergence in distribution.

\section{Background}

Suppose we have a model $f_{\th}$ parameterized by $\th$ in a parameter space $\Theta \subseteq \mathbb{R}^d$. Let $\{x_{i}\}_{i=1}^{n} \subset \mathcal X$ 
be a training set with $n$ data points on $\mathcal X$. Assume $\ell(x,f_{\th})$
is the negative log-likelihood of data point $x$ with model $f_{\th}$. A standard approach to estimate $\theta$ is 
maximum likelihood estimator (MLE), which minimizes 
the negative log-likelihood function (loss) over the training set
\vspace{-0mm}
\begin{eqnarray*}
\hat{\th}=\arg\min_{\th\in\Theta}\L(\th),~~~~~~~~
\L(\th)=\textstyle{\sum}_{i=1}^{n}\ell(x_{i},f_{\th})/n.
\end{eqnarray*}
Here the MLE $\hat{\theta}$ provides a point estimation without any information on the data uncertainty. Bootstrap is a simple and effective
frequentist method to quantify the uncertainty. The bootstrap loss is a randomly perturbed loss defined as
\[
\L_{\w}(\th)=\textstyle{\sum}_{i=1}^{n}w_{i}\ell(x_{i},f_{\th})/n,
\]
where $\w=[w_{1},...,w_{n}]^\top$ is a set of random weights of data points drawn from some distribution $\pi$. 
A typical choice of $\pi$ is 
the multinomial distribution with uniform probability, which corresponds to 
resampling on the training set with replacement. 
Given $\w$, one can calculate its associated bootstrap particle by minimizing the bootstrap loss:
\begin{align}\label{equ: bootstrap theta}
\hat{\th}_{\w}=\arg\min_{\th\in\Theta}\L_{\w}(\th).
\end{align}
Let $\rho_{\pi}$ be the distribution of $\hat\th_{\w}$ when $\w\sim\pi$. Bootstrap theory indicates that we can quantify the data uncertainty of $\th$ or any function $g(\th)$ using $\rho_{\pi}$. We call $\rho_\pi$ the ideal bootstrap distribution and it is the main object we want to approximate.

Denote $\delta_{\th}$ as the delta measure centered at $\th$. Standard bootstrap method approximates $\rho_{\pi}$ by the particle distribution $\hat{\rho}_{\pi}(\cdot)=\textstyle{\sum}_{j=1}^{m}\delta_{\hat{\th}_{\w_{j}}}(\cdot)/m$ formed by $m$ i.i.d. particles $\{\hat{\th}_{\w_{j}}\}_{j=1}^m$, which can be obtained by drawing $m$ i.i.d. weights $\{\w_{j}\}_{j=1}^{m}$ from $\pi$ and calculating each $\hat{\th}_{\w_{j}}$ based on (\ref{equ: bootstrap theta}).
%
However, for deep learning applications, as discussed in the introduction, storing and making inference using a large number $m$ of bootstrap particles can be quite expensive. On the other hand, if $m$ is small, $\hat{\rho}_{\pi}$ tends to be a poor approximation of $\rho_{\pi}$. In this paper, we aim to improve the approximation of the particle distribution when $m$ is small.
\section{Method} \label{sec: method}

Our idea is simple. Instead of using i.i.d. particles, in which the location of each particle is independent from that of the others, we try to actively optimize the location of each particle so that particles are diversified, better distributed and eventually providing a particle distribution with improved approximation accuracy. A natural way to achieve this goal is to explicitly optimize a set of points $\{\theta_j\}_{j=1}^m$ (called centroids) jointly such that the Wasserstein distance between $\rho_{\pi}$ and the induced particle distribution is minimized:
\begin{align} \label{equ:ideal_opt}
\{\th_{j}^{*},v_{j}^{*}\}_{j=1}^{m}=\hspace{-0.2cm}\underset{\th_{1},...,\th_{m}\in\Theta,[v_{1},...,v_{m}]\in\mathcal{C}^{m}}{\arg\min}\W_{2}\left[\sum_{j=1}^{m}v_{j}\delta_{\th_{j}},\rho_{\pi}\right]
\end{align}
Here we consider a Wasserstein distance $\W_2$ equipped with a data-dependent distance metric $||\cdot||_D$ that will be introduced later in (\ref{equ: taylor_intuition}). Note that here we also optimize the probability weights $\{v_j\}_{j=1}^m$ of the centroids. Finding the optimal centroids and probability weights can be decomposed into two steps: the centroid learning phase and the probability weights learning phase, based on the facts in (\ref{equ: w by dist},\ref{equ: v in opt w}).
\begin{align}
\label{equ: w by dist}
 & \W_{2}^{2}\bigg[{\textstyle \sum_{j\in[m]}}v_{j}^{*}\delta_{\th_{j}^{*}},\rho_{\pi}\bigg]=J_{\pi}(\{\th_{j}\}_{j=1}^{m}),\\
 \nonumber
\text{where\ } & J_{\pi}(\{\th_{j}\}_{j=1}^{m}):=\E_{\w\sim\pi}\left[\min_{j\in[m]}||\th_{j}-\hat{\th}_{\w}||_{D}^{2}\right]
\end{align}
Here (\ref{equ: w by dist}) implies that, to find the optimal particle distribution in (\ref{equ:ideal_opt}), we can start with the centroid learning phase where we only need to optimize the centroids. It can be achieved by minimizing
$
J_{\pi}(\{\th_{j}\}_{j=1}^{m})
$, which is the averaged distance of bootstrap particles to their closest centroid. After we obtain the optimal centroids, the optimal probability weights can be learned by (\ref{equ: v in opt w}):
\begin{align}
 \label{equ: v in opt w}
 & v_{j}^{*}=\tilde{v}_{j}^{*}/{\textstyle \sum_{s\in[m]}\tilde{v}_{s}^{*}},\\
 \nonumber
\text{where}\  & \tilde{v}_{j}^{*}=\P_{\w\sim\pi}\left(j=\arg\min_{j\in[m]}||\th_{j}^{*}-\hat{\th}_{\w}||_{D}^{2}\right)
\end{align}
Here $v_j^*$ is the proportion of bootstrap particles that are closest to the centroid $j$. We emphasize that the optimal solution to two-stage learning is guaranteed to be the global minimizer of the loss in (\ref{equ:ideal_opt}) (see Lemma 3.1 and 3.2 in \citet{NIPS2012_c54e7837}).

However, the key issue is that the losses in both (\ref{equ:ideal_opt}, \ref{equ: w by dist}) can not be computed in practice, as they require us to access $\rho_\pi$ (i.e., obtain $\hat{\th}_{\w}$ first in order to calculate the loss). To handle this issue, we seek an easy-to-compute surrogate loss. Our idea is based on the following observation. Assuming the size of training data is large, which is usually the case in deep learning, we can expect that $\th_{\w}$ will be centered around a small region\footnote{This can be formally characterized by central limit theorem as discussed in Section \ref{sec: theory}.}. It implies that we should search the centroid in this small region. Notice that when $\th$ is close to $\hat{\th}_{\w}$, based on Taylor approximation, we have
\begin{align} \label{equ: taylor_intuition}
\nonumber
\L_{\w}(\th) & \approx\L_{\w}(\hat{\th}_{\w})+\nabla_{\th}^{\top}\L_{\w}(\hat{\th}_{\w})(\th-\hat{\th}_{\w})\\
\nonumber
 & +\frac{1}{2}(\th-\hat{\th}_{\w})^{\top}\nabla_{\th}^{2}\L_{\w}(\hat{\th}_{\w})(\th-\hat{\th}_{\w})\\
 & \approx\L_{\infty}(\th_{0})+||\th-\hat{\th}_{\w}||_{D}^{2},
\end{align}
where $\left\Vert V\right\Vert _{D}^{2}:=V^{\top}\nabla_{\th}^{2}\L_{\infty}(\th_{0})V$.
Here $\L_\infty(\th) := \E_x \ell(x, f_\th)$ denotes the population loss; $\th_0$ is the minimizer of $\L_\infty(\th)$. In (\ref{equ: taylor_intuition}), we use the facts\footnote{We defer the detailed analysis to Section \ref{sec: theory}.} that $\nabla_{\th}^{\top}\L_{\w}(\hat{\th}_{\w})=0$;
and with large training set, the empirical distribution $\textstyle{\sum}_{i=1}^n \delta_{x_i}/n$ well approximates the whole data population, and hence the bootstrap resampling distribution, i.e., $\textstyle{\sum}_{i=1}^n w_i\delta_{x_i}/n$ on the empirical distribution also well approximates the whole data population. This implies that $\L_{\w}(\cdot)\approx\L_{\infty}(\cdot)$ and $\nabla_{\th}^{2}\L_{\w}(\cdot)\approx\nabla_{\th}^{2}\L_{\infty}(\cdot)$. As the loss are close to each other, their minimizers are also close $\hat{\th}_{\w}\approx\th_{0}$. Since $\L_{\infty}(\th_{0})$ is some (unknown) constant independent with $\th$, we can replace the $||\th_j-\hat{\th}_{\w}||^2_D$ in (\ref{equ: w by dist}) by $\L_{\w}(\th_j)$ as it only adds some constant into the loss.

Intuitively, we can expect that the centroid closest to $\hat{\th}_{\w}$ is the one that gives the smallest loss on $\L_{\w}$. It motivates us to learn the centroids via the modified centroid learning phase:
\begin{align}\label{equ:opt}
\hspace{-0.3cm}
\{\th_{j}^{*}\}_{j=1}^{m}=\arg\min_{\th_1,...,\th_m\in\Theta}\E_{\w\sim\pi}\left[\textstyle{\min}_{j\in[m]}\L_{\w}(\th_{j})\right].
\end{align}
Similarly, the optimal probability weights can be learned via the modified weight learning phase:
\begin{align} \label{equ:opt_v}
 & v_{j}^{*}=\tilde{v}_{j}^{*}/{\textstyle \sum_{s\in[m]}\tilde{v}_{s}^{*}},\ \\
 \nonumber
\text{where}\  & \tilde{v}_{j}^{*}=\P_{\w\sim\pi}\left(j\in\arg{\textstyle \min}_{j\in[m]}\L_{\w}(\th_{j}^{*})\right)\end{align}
We note that here we slightly abuse the notation of $\th_j^*$ and $v^*_j$ in (\ref{equ: w by dist},\ref{equ: v in opt w}) and (\ref{equ:opt},\ref{equ:opt_v}) for simplification. In the later context, $\th_j^*$ and $v^*_j$ are used based on their definitions in (\ref{equ:opt},\ref{equ:opt_v}).

\paragraph{Connection to K-means}
By viewing the target distribution as a set of particles that we want to cluster, in K-means clustering, each centroid (i.e., K-means center) represents one of the K disjoint groups\footnote{i.e. the regions separated by the dashed lines in the right plot of Figure \ref{fig: illustration}.} of particles, which is formed by assigning each particle in the whole set to the closest centroid among all the K centroids. K-means learns the optimal K centroids in the way that they can best approximate the whole set. The `closeness' for assigning the particles is measured by the distance between the two points. As pointed out by \citet{NIPS2012_c54e7837}, K-means essentially searches the optimal particle distribution formed by the K centroids that minimizes its Wasserstein distance to the target distribution. Our centroid approximation idea follows the same fashion of clustering but our key innovation is to measure the `closeness' by examining the bootstrap loss of the centroids so that we can still learn the optimal centroids without obtaining the i.i.d. bootstrap particles first. We also point out that, while we share the same objective as K-means, the optimization algorithms differ. The Expectation-Maximization type of algorithm used by K-means is not applicable to our scenario.

{\color{black} \paragraph{Comparing with Other Particle Improving Approach}
Intuitively, from a high level abstracted perspective, we provide an approach to use K-means type of idea to improve the particle quality \emph{without accessing to the true target distribution}. This is the key differentiator of this work to other approaches that improve the particle quality, as they all require to access the target distribution. For example \citet{claici2018wasserstein} requires that sampling from target distribution is cheap and easy. \citet{chen2012super,chen2018stein,campbell2019sparse} need to access the logarithm of the probability density function of the target distribution. In our problem, neither sampling from the target distribution is cheap nor the logarithm of the probability density function is available, making those approaches no more applicable.} In concurrent work \citep{gong2021argmax} the idea to improve particle is applied to multi-domain learning problems. Compared with \citet{gong2021argmax}, we focus on bootstrap inference that allows us to do fine-grained theoretical analysis and fast computation.

\paragraph{Comparing with m-out-of-n/bag-of-little Bootstrap}
The m-out-of-n bootstrap \citep{bickel2012resampling} and the bag-of-little bootstrap \citep{kleiner2014scalable} are designed to reduce the computational cost with the subsampling techniques in the big data settings (large $n$). Our method and m-out-of-n bootstrap/bag-of-little bootstraps are working towards two orthogonal directions of improving the scalability of bootstrap. m-out-of-n bootstrap/bag-of-little bootstraps aim to decrease the training cost when the size of the dataset is large while our paper improves the approximation accuracy when a limited number of bootstrap particles are allowed at inference is small (small $m$).

\subsection{Training}
The optimization of (\ref{equ:opt}) can be solved by gradient descent. Suppose $\th_{j}^{*}(t)$
is the $j$-th centroid at iteration $t$. We initialize $\{\th^*_j(0)\}_{j=1}^{m}$ by sampling from $\rho_\pi$ and at iteration $t$, we update $\th_{t}^{*}$ by applying the gradient descent on the loss in (\ref{equ:opt}), which yields 
\begin{align} \label{upadte: centroid}
\begin{split} 
& \hspace{-0.32cm} \th_{j}^{*}(t+1)\leftarrow\th_{j}^{*}(t)-
\epsilon_{t}
g(\theta_j^*(t)),  \\
& 
\hspace{-0.32cm}g (\theta_j^*(t)) = {\E_{\w\sim\pi}\left[
\mathbb{I}\{{j\in u_{\w}(t)}\}
\nabla_{\th}\L_{\w}(\th_{j}^{*}(t))\right]}/{v_j^*(t)},
\end{split} \end{align}
where we define the index of the closest centroid to particle $\hat{\th}_{\w}$ as $u_{\w}(t) = 
\arg\min_{j\in[m]}\L_{\w}(\th_{j}^{*}(t))$ and  $v_j^*(t) = \P_{\w\sim\pi}\left( 
j\in u_{\w}(t)\right)$ denotes the probability that centroid $j$ is the one that gives the lowest bootstrap loss. 
The denominator $v^*_j(t)$ in $g(\th_j^*(t))$ is optional. However, notice that the magnitude of numerator in $g(\th_k^*(t))$ decays with larger $m$, which might require an adjustment of the learning rate when $m$ changes. This adjustment can be avoided by rescaling with $v^*_k(t)$.

We note that $\{\th^*_j(0)\}_{j=1}^{m}$ is just $m$ i.i.d. bootstrap particles which is not optimal for approximation and our algorithm can be viewed as an approach for refining the $m$ particles by solving (\ref{equ:opt}). In practice, we find that we can simply use random initialization (e.g., draw $\th$ from some Gaussian distribution) instead.

\paragraph{Centroid Degeneration Phenomenon}
Naively applying the updating rule (\ref{upadte: centroid}) may cause a degeneration phenomenon: When a centroid happens to give considerably worse performance than others, which can be caused by the stochasticity of gradient or worse initialization, the performance of this centroid will remain considerably worse throughout the optimization. The reason is simple. As this centroid (e.g. $\th_j^*(t)$) gives a considerably worse performance, the probability that it gives the lowest bootstrap loss, i.e., $v_j^*(t)$, is small. As a consequence, the gradient that updates this centroid is only based on aggregating information from a small low-density region of $\pi$ and hence can be unstable and further degrades this centroid. Note that this mechanism is self-reinforced since when this centroid cannot be effectively improved in the current iteration, it faces the same issue in the next one. As a result, this centroid is always significantly worse than the others.

We call this undesirable phenomenon \emph{centroid degeneration} and we want to prevent this phenomenon because when it happens, we have a centroid that is not representative and contributes less to approximating $\rho_\pi$. We solve this issue with a simple solution and here is the intuition. The reason that a centroid degenerates lies in that this centroid is far from the good region where it gives a good performance. And when this happens, we should push the centroid to move towards this good region, which can be achieved by using the common gradient over the whole training data. Specifically, we define a threshold $\gamma$, indicating centroid $j$ is degenerated if $v_j^*(t)\le\gamma$. And when it happens, we update using the common gradient over the whole data:
\begin{align}\label{upadte: centroid_degenerated}
\th_{j}^{*}(t+1)\leftarrow\th_{j}^{*}(t)-\epsilon_{t}\nabla_{\th}\L(\th_{j}^{*}(t)).
\end{align}
In section \ref{sec: theory}, we give a theoretical analysis on why this modification is important and is able to solve the centroid degeneration issue.

\paragraph{Practical Algorithm}
In practice, we estimate the gradient by replacing the expectation over $\w\sim\pi$ in (\ref{upadte: centroid}) with averaging over $M$ i.i.d. Monte Carlo samples $\{\w_{h}\}_{h=1}^{M}$ drawn from $\pi$: 
\begin{align} \label{equ: g_hat}
\hat{g}(\th_{j}^{*}(t))=\frac{\sum_{h=1}^{M}\left[\mathbb{I}\{j\in u_{\w_{h}}(t)\}\nabla_{\th}\L_{\w_{h}}(\th_{j}^{*}(t))\right]}{\sum_{h=1}^{M}\mathbb{I}\{j\in u_{\w_{h}}(t)\}}.
\end{align}

Now it remains to compute $\w_{h}$ and $\L_{\w_{h}}$ for all $h\in[M]$, which can be done very cheaply by firstly compute
\begin{align}\label{equ: datawise loss}
\LL(\th_{j}^{*}(t))=[\ell(x_{1},f_{\th_{j}^{*}(t)}),...,\ell(x_{n},f_{\th_{j}^{*}(t)})]^{\top}\in\mathbb{R}^{n}
\end{align}
and then compute $\L_{\w_{h}}=\w_{h}^{\top}\LL(\th_{j}^{*}(t))$ and 
\begin{align} \label{equ: uwh}
u_{\w_{h}}(t)=\arg\min_{j\in[m]}\w_{h}^{\top}\LL(\th_{j}^{*}(t)).
\end{align}
Taking the modified updating rule introduced to prevent the centroid degeneration phenomenon into account, we update $\theta_j^*(t)$ by $\th_{j}^{*}(t+1)\leftarrow\th_{j}^{*}(t)-\epsilon_t\phi(\th_{j}^{*}(t))$, where
\begin{align}
\label{update: practical}
\phi(\th_{j}^{*})=\begin{cases}
\hat{g}(\th_{j}^{*}(t)) & \hspace{-0.2cm} \text{if}\sum_{h\in[M]}\mathbb{I}\{u_{\w_{h}}(t)=j\}/M>\gamma\\
\nabla_{\th}\L(\th_{j}^{*}(t)) & \hspace{-0.2cm} \text{otherwise.}
\end{cases}
\end{align}
Algorithm \ref{algo_ideal} summarizes the whole procedure. Note that as $\LL(\th_{j}^{*}(t))$ can be reused for computing all $\L_{\w_{h}}$ and $u_{\w_{h}}(t)$, $h\in[M]$. The computation overhead is hence very small ($O(nM)$ matrix multiplication for each centroid).

In practical implementation, as $u_{\w_{h}}(t)$ do not change much within a few iterations, we can update $u_{\w_{h}}(t)$ every a few iterations (e.g., every epoch). We can also replace the $\nabla_{\th}\L_{\w_{h}}(\th_{j}^{*}(t))$ or $\nabla_{\th}\L(\th_{j}^{*}(t))$ in (\ref{update: practical}) using a mini-batch of data instead of the whole data, which leads to a stochastic gradient version of our algorithm. Due to space limit, we summarize the algorithm using stochastic gradient in Algorithm \ref{algo} in Appendix \ref{apx_sec: alpo}.

\begin{algorithm}[t]
\begin{algorithmic}[1]
\STATE{Initialize $\th^*_j(0)$, $j\in[m]$ by i.i.d. sampling from $\rho_\pi$ or other distribution such as Gaussian.}
\FOR{$t \in \text{iterations}$}
    \STATE{$\forall j\in[m]$, calculate $\LL(\th^*_j(t))$ defined in (\ref{equ: datawise loss})}
    \STATE{Sample $\{\w_h\}_{h=1}^M$, i.i.d. from $\pi$.}
    \STATE{$\forall h \in [M]$ and $j \in [m]$, calculate $\LL_{\w_h}(\th^*_j(t))=\w_h^T\LL(\th_j^*(t))$.}
    \STATE{$\forall h \in [M]$, calculate $u_{\w_h}$ defined in (\ref{equ: uwh}).}
    \STATE{$\forall j \in [m]$, update $\th^*_j$ by (\ref{update: practical}).}
\ENDFOR
\end{algorithmic}\caption{Ideal algorithm for centroid approximation with full-batch gradient and $\w_h$ updated every iteration.} \label{algo_ideal}
\end{algorithm}

\section{Theory} \label{sec: theory}
Recall that, as discussed in (\ref{equ: taylor_intuition}), our approach relies on the intuition that bootstrap particles are nested in a small region so that we can approximate the distance between the centroid and a bootstrap particle by the bootstrap loss of that centroid. The main goal of this section is to give a formal theoretical justification of this intuition.

Before we proceed, we clarify several important setups for establishing and interpreting the theoretical result. As discussed in the introduction, we are mainly interested in the scenerio that the number of available particles/centroids $m$ is small while the number of training data $n$ is large, which motivates us to establish theoretical result in the region of small $m$ and large $n$. This is significantly different from conventional asymptotic analysis in which we aim to show the behavior when $m \to \infty$. {\color{black} We assume that the parameter dimension $d$ is fixed and does not scale with $n$.}

\emph{We are mainly interested in characterizing the approximation of the proposed loss in (\ref{equ:opt}) to the ideal loss in (\ref{equ: w by dist}), given any small and fixed number $m$ of centroids when $n\to\infty$. This justifies why the proposed centroid approximation method can be viewed as minimizing the Wasserstein distance between the particle distribution $\rho_\pi^*$ and the target bootstrap distribution $\rho_\pi$.}

For simplicity, we build our analysis assuming the ideal update rule (\ref{upadte: centroid},\ref{upadte: centroid_degenerated}) is used. We start with the following main assumptions.

\begin{assumption}[Smoothness and boundedness] \label{assumption: bound}
Assume that the following quantities are upper bounded by some constant $c<\infty$:
\begin{align*}
 & 1.\ \sup_{\th_{1},\th_{2}\in\Theta}\sup_{x\in\mathcal{X}}\frac{||\nabla_{\th}^{2}\ell(x,f_{\th_{1}})-\nabla_{\th}^{2}\ell(x,f_{\th_{2}})||}{||\th_{1}-\th_{2}||};2.\ \sup_{\th\in\Theta}\left\Vert \th\right\Vert;  \\
 & 3.\ \max_{i,j,k\in[d]}\sup_{\th\in\Theta,x\in\mathcal{X}}\frac{\partial^{3}\ell(x,f_{\th})}{\partial_{i}\th_{i}\partial\th_{j}\partial\th_{k}};4.\ \sup_{x\in\mathcal{X},\th\in\Theta}\left\Vert \nabla_{\th}^{2}\ell(x,f_{\th})\right\Vert .
\end{align*}
\end{assumption}
Assumption \ref{assumption: bound} is a standard regularity condition on the boundness and smoothness of the problem.
\begin{assumption}[Asymptotic normality] \label{assumption: asy-normal}
Assume $\sqrt{n}\left(\hat{\th}_{w}-\hat{\th}\right)\overset{d}{\to}\N\left(0,A\right)$ and $\sqrt{n}\left(\hat{\th}-\th_{0}\right)\overset{d}{\to}\N\left(0,A\right)$ as $n\to \infty$, where $A$ is a positive-definite matrix with the largest eigenvalue bounded.
\end{assumption}
Assumption \ref{assumption: asy-normal} is a higher level assumption on the asymptotic normality of the estimators. Such result is classic and can be derived with some weak and technical regularity conditions. See examples in \citet{chatterjee2005generalized,cheng2010bootstrap}.
\begin{assumption}[On the global minimizer] \label{assumption: min_eigen}
Suppose that $\lambda_{\min}\left(\nabla_{\th}^{2}\L_{\infty}(\th_{0})\right) > 0$.
\end{assumption}
Assumption \ref{assumption: min_eigen} is also standard showing the locally strongly convexity of the loss around the truth $\th_0$.
\begin{assumption}[On the learning rate] \label{assumption: lr}
Suppose that $\max_t \epsilon_t = O(n^{-1})$.
\end{assumption}
Assumption \ref{assumption: lr} assumes that the learning rate of the algorithm is sufficiently small such that its induced discretization error is not the dominating term.

\begin{table*}[t]
\centering{}%
\scalebox{1.}{
\begin{tabular}{c|c|c|cccc}
\toprule
\multicolumn{1}{c}{} & \multicolumn{2}{c|}{} & $m=20$ & $m=50$ & $m=100$ & $m=200$\tabularnewline
\hline 
\multirow{6}{*}{$\alpha=0.9$} & \multirow{2}{*}{Normal} & Bootstrap & $0.029\pm0.010$ & $0.031\pm0.011$ & $0.021\pm0.010$ & $0.017\pm0.010$\tabularnewline
 &  & Centroid & $\textbf{0.027}\pm\textbf{0.010}$ & $\textbf{0.001}\pm\textbf{0.009}$ & $\textbf{0.012}\pm\textbf{0.010}$ & $\textbf{0.016}\pm\textbf{0.010}$\tabularnewline
\cline{2-7} \cline{3-7} \cline{4-7} \cline{5-7} \cline{6-7} \cline{7-7} 
 & \multirow{2}{*}{Percentile} & Bootstrap & $0.101\pm0.013$ & $0.036\pm0.011$ & $0.021\pm0.010$ & $\textbf{0.014}\pm\textbf{0.010}$\tabularnewline
 &  & Centroid & $\textbf{0.081}\pm\textbf{0.012}$ & $\textbf{0.021}\pm\textbf{0.010}$ & $\textbf{0.020}\pm\textbf{0.010}$ & $0.015\pm0.010$\tabularnewline
\cline{2-7} \cline{3-7} \cline{4-7} \cline{5-7} \cline{6-7} \cline{7-7} 
 & \multirow{2}{*}{Pivotal} & Bootstrap & $0.106\pm0.013$ & $0.045\pm0.011$ & $0.025\pm0.010$ & $0.023\pm0.010$\tabularnewline
 &  & Centroid & $\textbf{0.046}\pm\textbf{0.011}$ & $\textbf{0.013}\pm\textbf{0.009}$ & $\textbf{0.011}\pm\textbf{0.010}$ & $\textbf{0.020}\pm\textbf{0.010}$\tabularnewline
\bottomrule
\end{tabular}
}\caption{Centroid approximation for confidence interval. The numbers in the table represent $|\alpha-\hat \alpha|$, where $\hat \alpha$ is the estimated coverage probability. The errors bar is the standard deviation.}
\label{table: ci}
\end{table*}

The key challenge of our analysis is to show that our dynamics is $\mathcal{B}(\th_0,r)$-stable (defined below in Definition \ref{def: stable}) for some small $r$, saying that $\{\th_j^*(t)\}_{j=1}^m$ stay in a small region that is close to $\th_0$ \emph{for any iteration t}. Combined with the property\footnote{This is implied by the asymptotic normality in assumption \ref{assumption: asy-normal}.} that $\hat{\th}_{\w}$ are also close to $\th_0$, the centroids and the bootstrap particles are close to each other and thus our approximation in (\ref{equ: taylor_intuition}) holds for all $t\ge0$. In this way, optimizing the centroids by minimizing our loss is almost equivalent to optimizing the centroids by minimizing the Wasserstein distance. 

\begin{definition}[$\mathcal{B}(\theta,r)$-stable] \label{def: stable}
Given some $\th\in\Theta$ and $r\ge0$, we say our dynamics is $\mathcal{B}(\theta,r)$-stable if $\forall t\ge 0$ and $\forall j\in[m]$, $\th_{j}^{*}(t)\in\mathcal{B}(\th,r)$, where $\mathcal{B}(\th,r):=\{\th':\left\Vert \th'-\th\right\Vert \le r,\ \th'\in\Theta\}$ is the ball with radius $r$ centered at $\th$.
\end{definition}

The key intuition to establish such $\mathcal{B}(\th_0,r)$-stable result is to characterize that our optimization dynamics is implicitly \emph{self-controlled}: when some centroid approaches the boundary of $\mathcal{B}(\th_{0},r)$, the updating mechanism automatically start to push the centroid to move towards the center of the region. Thus, if all the centroids are within $\mathcal{B}(\th_{0},r)$ at initialization, they will alway stay in this region.

Thanks to assumption \ref{assumption: asy-normal}, \ref{assumption: min_eigen}, when the dataset is large, the landscape of our loss is locally strongly convex around $\th_{0}$. When a centroid $j$ is at the boundary of $\mathcal{B}(\th_{0},r)$, it has $v_j^*(t)<\gamma$ and thus the updating direction is the gradient of loss $\L$. By the convexity, such gradient will push the centroid move towards the center of $\mathcal{B}(\th_{0},r)$ where the empirical minimizer locates at. On the other hand, for centroid $j$ with $v_j^*(t)\ge\gamma$, its updating direction aggregates information from sufficient data point and thus behaves similarly to that of the common gradient, pushing centroid to move towards the center with the centroid is not close to the center.

\begin{theorem} \label{thm: stable_main}
Under Assumptions 1-4 and suppose that we initialize $\th^*_j(0)$, $j\in[m]$ by sampling from $\rho_\pi$, given any $m<\infty$ and $\gamma>0$, {\color{black} when $n$ is sufficiently large,} we have 
\[
\max_{j\in[m]}\sup_{t\ge0}\left\Vert \th_{j}^{*}(t)-\th_{0}\right\Vert =O_{p}(\sqrt{(\log n)/n}).
\]
Here the probability is taken w.r.t. training data.
\end{theorem}
Theorem \ref{thm: stable_main} implies our dynamics is $\mathcal{B}(\th_{0},r_{n})$-stable with $r_n=O(\sqrt{\log n/n})$. The condition that $\th^*_j(0)\sim\rho_\pi$ i.i.d. can be replaced by the condition that $\th^*_j(0)$ is sufficiently close to $\th_0$. We need such condition as we uniformly bound the distance between $\th^*_j(t)$ and $\th_0$ at any iteration including the first one. Theorem \ref{thm: stable_main} {\color{black}implies that the approximation stated in (\ref{equ: taylor_intuition}) holds with high probability and hence} the proposed loss in (\ref{equ:opt}) is `almost as good as' the ideal loss in (\ref{equ: w by dist}).

\begin{theorem} \label{thm: good_surrogate}
Under the same assumptions as Theorem \ref{thm: stable_main}, given any $m<\infty$ and $\gamma>0$, {\color{black} when $n$ is sufficiently large}, we have
\begin{scriptsize}
\begin{align*}
 & \sup_{t\ge0}\left|\E_{\w\sim\pi}[\min_{j\in[m]}\L_{\w}(\th_{j}^{*}(t))]-B-\E_{\w\sim\pi}[\min_{j\in[m]}||\th_{j}^{*}(t)-\hat{\th}_{\w}||_{D}^{2}]/2\right|\\
 & = O_{p}(\sqrt{\log n/n^{3/2}})
\end{align*}
\end{scriptsize}
Here the probability is taken w.r.t. training data and $B$ is some constant independent from $\th_j^*(t)$ for any $t\ge0$ and $j\in[m]$.
\end{theorem}

\paragraph{Asymptotics when $m$ also grows}
Although our main interest is the asymptotics with a small, fixed $m$ and growing $n$, we discuss here on asymptotics when $m$ also grows. As shown in Section \ref{sec: method} and introduction, our method can be viewed as an `approximated' K-means on the target distribution. From Theorem 5.2 in \citet{NIPS2012_c54e7837}, the particle distribution formed by the optimal centroids learned by K-means gives improved $O(m^{-1/d})$ convergence to any general target distribution in terms of Wasserstein distance, where $d$ is data dimension. In comparison, the particle distribution of i.i.d. sample only gives $O(m^{-1/(2d+4)})$ from Theorem 5.1 in \citet{NIPS2012_c54e7837}. This implies that our approach potentially also has such a rate improvement. Note that the results in \citet{NIPS2012_c54e7837} are for general target distribution without any $n$ involves. To rigorously establish the large $m$ asymptotic result for our problem, we need to study the joint limit of $n$ and $m$. This is indeed very non-trivial: as discussed in \citet{weed2019sharp} (i.e. Proposition 14), when $n\to\infty$, the target distribution $\rho_{\pi}$ becomes a sharp Gaussian and the convergence rate of i.i.d. bootstrap particles will gradually improve to $O(m^{-1/2})$ (in a way that depends on $n$). It implies that when $n\gg m\to\infty$, our improvement may become only constant level. We find establishing such a theory is out the scope of this conference paper and leave it as future work.

\section{Experiment}
We aim to answer the following questions:

\textbf{Q1:} Whether the proposed objective effectively approximates the Wasserstein distance between the particle distribution and the target distribution? (Yes)

\textbf{Q2:}  Whether our approach improves the quality of the particle distribution when only a limited number of particles/centroids are allowed at inference time? (Yes)

\textbf{Q3:} While our main goal is not to decrease the training cost but improve the quality of bootstrap particle distribution, as discussed in Section \ref{sec: method}, our method actually only introduces a little training overhead, which is another advantage of our method. Our third question is whether our approach truly gives small training overhead in practice. (Yes)

\textbf{Q4:} Whether the modification in (\ref{update: practical}) improves the learning by overcoming the centroid degeneration issue? (Yes)

To answer \textbf{Q1}, we show that when applied to confidence interval estimation (section \ref{sec: confidence interval}), compared with naive Bootstrap, the particles learned by our centroid approximation approach gives significantly smaller Wasserstain distance to the target.

To answer \textbf{Q2}, we apply our approach to various application including confidence interval estimation (section \ref{sec: confidence interval}), contextual bandit (section \ref{sec:contextualbandit}), bootstrap DQN (section \ref{sec:bootstrapdqn}) and bagging (Appendix \ref{apx_sec: bagging}, due to space limit) and we demonstrate that with different choice of $m$ ($m$ is small), our approach consistently gives improvement over the bootstrap baseline.

To answer \textbf{Q3}, we compared the training time between naive bootstrap and our centroid approximation for various applications and show that the training overhead of our approach is very small. Due to the space limit, we refer readers to Appendix \ref{apx_sec: overhead} for details.

To answer \textbf{Q4}, we conduct ablation study on difference choice of $\gamma$ showing the improvement of applying the modification in (\ref{update: practical}). Due to the space limit, we refer readers to Appendix \ref{apx_sec: ablation} for details.

Code is available at \url{https://github.com/lushleaf/centroid_approximation}.

\subsection{Bootstrap Confidence Interval} \label{sec: confidence interval}

\begin{figure}[t]
  \begin{minipage}[c]{0.25\textwidth}
    \includegraphics[width=\textwidth]{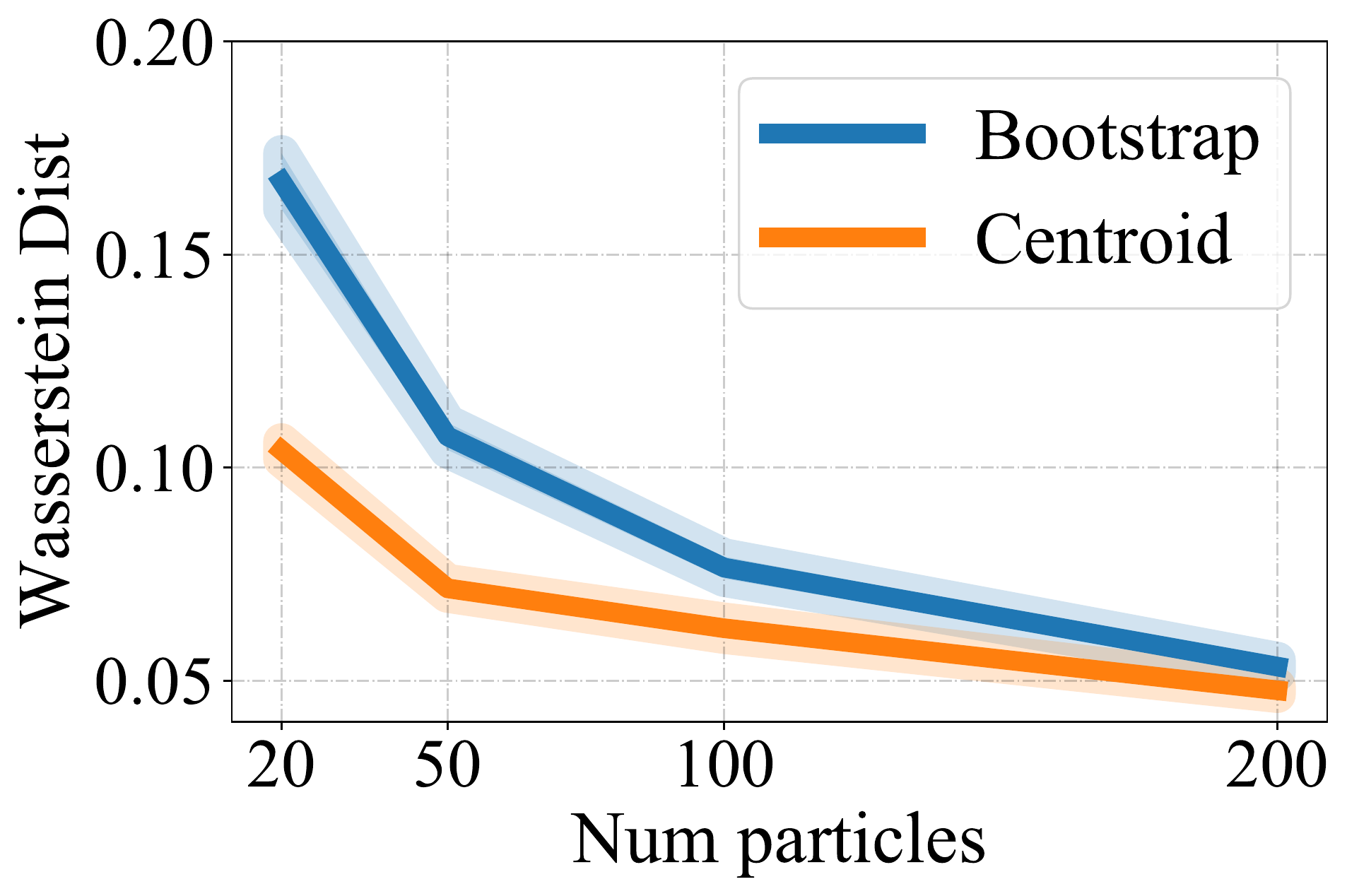}
    \vspace{-0.6cm}
  \end{minipage}
  \begin{minipage}[c]{0.19\textwidth}
    \vspace{-1cm}
    \caption{
       Wasserstein distance between the particle distribution and the true bootstrap distribution w.r.t. the number of particles.
    } \label{fig: wasserstein}
    \vspace{-0.6cm}
  \end{minipage}
\end{figure}

We start with a classic application of bootstrap: confidence interval estimation for linear model with parameter $\th$. Fix confidence level $\alpha$, we consider three ways to construct (two-sided) bootstrap confidence interval of $\th$: the Normal interval, the percentile interval and the pivotal interval. And we test $m=20, 50, 100, 200$. For all experiments, we repeat with 1000 independent random trials.  
We consider the standard bootstrap as baseline. Detailed experimental setup are included in Appendix \ref{apx_sec: ci}.

Figure \ref{fig: wasserstein} shows the Wasserstein distance between the true target distribution $\rho_\pi$ and the empirical distributions obtained by (a) i.i.d. sampling $\hat{\rho}_\pi$, (b) the proposed centroid approximation $\rho^*_\pi$. The centroid approximation significantly reduces the Wasserstein distance by a large margin. 
We then compare the quality of obtained confidence intervals, which is measured by the difference between the estimated coverage probability and the true confidence level, i.e., $|\hat{\alpha}-\alpha|$ (the lower the better). 
Here we only consider confidence intervals of the first coordinate of $\theta$: $\theta_1$. Table \ref{table: ci} summarizes the result with $\alpha=0.9$. We see that using more particles is generally able to improve the constructed confidence intervals. We also compare with two variants of standard bootstrap: Bayesian bootstrap \citep{rubin1981bayesian} and residual bootstrap \citep{efron1992bootstrap}. And we consider varying $\alpha=0.8, 0.95$. These results are included in Appendix \ref{apx_sec: ci}.

\subsection{Centroid Approximation for Bootstrap Method in Contextual Bandit} \label{sec:contextualbandit}
\begin{table*}[t]
\begin{centering}
\scalebox{1}{
\begin{tabular}{c|c|llll}
\toprule
\multicolumn{1}{c}{} &   & \multicolumn{1}{c}{$m=3$} & \multicolumn{1}{c}{$m=4$} & \multicolumn{1}{c}{$m=5$} & \multicolumn{1}{c}{$m=10$}\tabularnewline
\hline 
\multirow{2}{*}{Mushroom} & Bootstrap & $3282.1\pm72.8$ & $3307.9\pm69.2$ & $3311.6\pm79.3$ & $3397.4\pm51.4$\tabularnewline
 & Centroid & $\textbf{3702.7}\pm\textbf{89.8}$ & $\textbf{3723.1}\pm\textbf{78.7}$ & $\textbf{3799.6}\pm\textbf{84.2}$ & $\textbf{3796.9}\pm\textbf{36.1}$\tabularnewline
\hline 
\multirow{2}{*}{Statlog} & Bootstrap & $1864.3\pm6.4$ & $1869.2\pm5.2$ & $1877.2\pm4.1$ & $1877.0\pm2.7$\tabularnewline
 & Centroid & $\textbf{1893.6}\pm\textbf{6.0}$ & $\textbf{1892.6}\pm\textbf{3.6}$ & $\textbf{1891.3}\pm\textbf{3.5}$ & $\textbf{1892.6}\pm\textbf{2.8}$\tabularnewline
\hline 
\multirow{2}{*}{Financial} & Bootstrap & $2255.8\pm58.4$ & $2265.4\pm58.2$ & $2269.3\pm56.4$ & $2281.4\pm56.6$\tabularnewline
 & Centroid & $\textbf{2313.3}\pm\textbf{56.4}$ & $\textbf{2315.3}\pm\textbf{56.7}$ & $\textbf{2323.9}\pm\textbf{56.7}$ & $\textbf{2325.5}\pm\textbf{56.0}$\tabularnewline
\bottomrule
\end{tabular}
}
\par\end{centering}
\caption{Results on the contextual bandit experiment. The numbers in the table represent the averaged reward with its standard deviation. Bolded value indicates that the better approach is statistical significant using matched pair t-test with p value less than 0.05.} \label{table: contextual}
\end{table*}

Contextual bandit is a classic task in sequential decision making, in which accurately quantifying the model uncertainty is important in order to achieve good exploration-exploitation trade-off. As shown in \citet{riquelme2018deep}, tracking the model uncertainty using bootstrap is a strong method for contextual bandit. However, it is costly to maintain a large number of bootstrap models and thus the number of models is typically within 10 \citep{osband2016deep}.
We find that applying the proposed centroid approximation here can significantly improve the performance. \citet{riquelme2018deep} uses $m=3$ bootstrap models and we give a more comprehensive evaluation with $m=3, 4, 5 ,10$. We consider three datasets: Mushroom, Statlog and Financial. We set $\gamma = 0.5/m$. We randomly generate 20 different context sequences, apply all the methods and report the averaged cumulative reward and its standard deviation. Table \ref{table: contextual} summarizes the result and note that a large part of variance can be explained by different context sequences. All results in Table \ref{table: contextual} are statistically significant under significant level 5\% using matched pair t-test. Table \ref{table: contextual} shows that using more bootstrap models generally improves the accumulated reward. And when using the same number of models, the proposed centroid approximation method consistently improves over standard bootstrap method. We refer readers to appendix \ref{apx_sec: cb} for more information on the background and experiment. The detailed executed algorithm is summarized in Algorithm \ref{algo_bootstrap_ci} in Appendix \ref{apx_sec: cb}.

\begin{figure}
  \begin{minipage}[c]{0.5\textwidth}
    \includegraphics[width=0.49\textwidth]{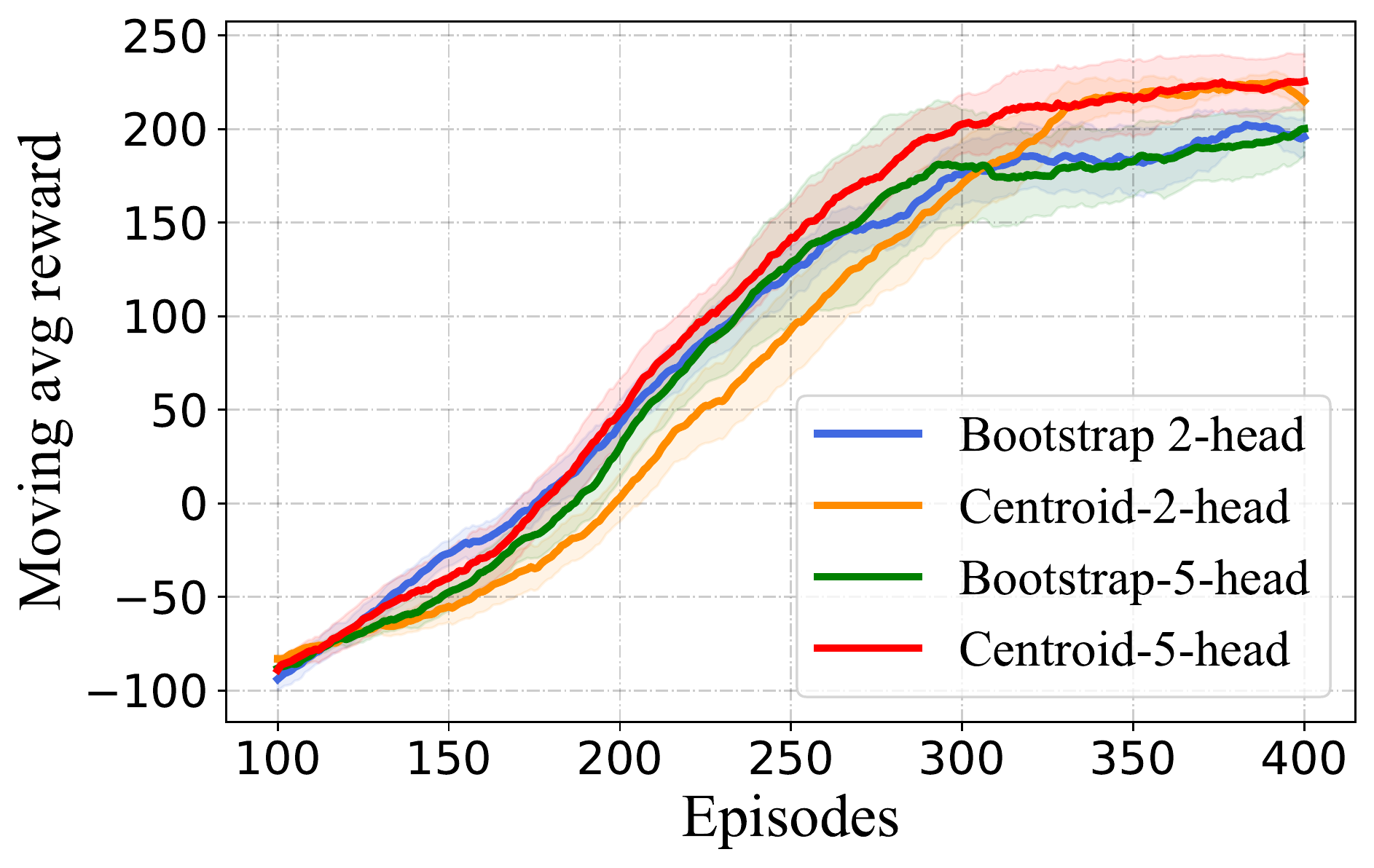}
    \includegraphics[width=0.49\textwidth]{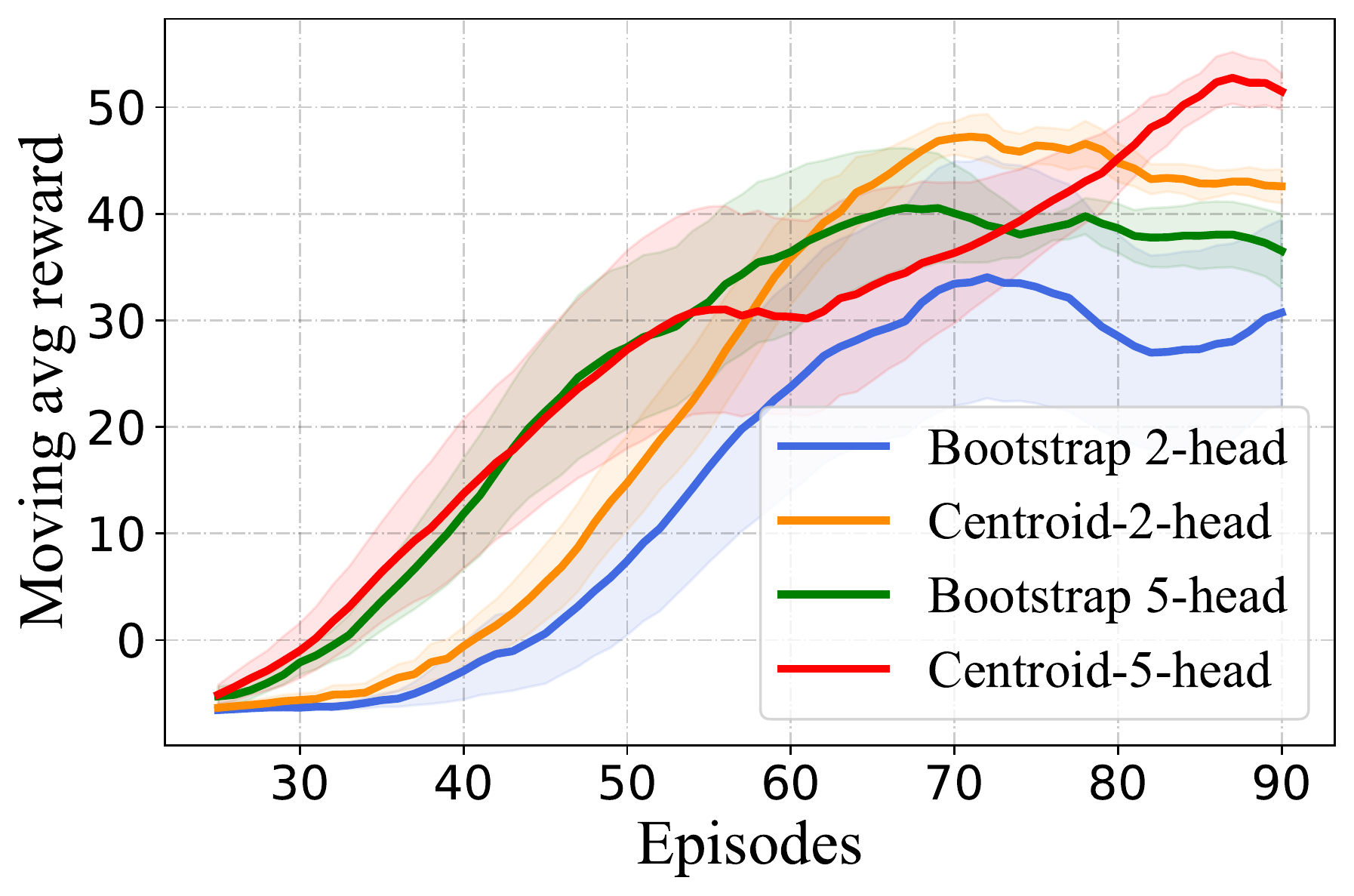}
  \end{minipage}
  \caption{
      Wasserstein distance between the particle distribution and the true bootstrap distribution w.r.t. the number of particles. Left: result for LunarLander; Right: result for Catcher.
    } \label{fig: dqn}
\end{figure}

\subsection{Centroid Approximation for Bootstrap DQN} \label{sec:bootstrapdqn}
{\color{black}``}Efficient exploration is a major challenge for reinforcement learning (RL). Common dithering strategies such as $\epsilon$-greedy do not carry out temporally-extended exploration, which leads to exponentially larger data requirements{\color{black}"} \citep{osband2016deep}. To tackle this issue, \citet{osband2016deep} proposes the Bootstrapped Deep Q-Network (DQN). We apply our centroid approximation to improve Bootstrapped DQN. We consider $m=2,\ 5$ and similar to the experimental setting in contextual bandit, we set $\gamma = 0.5/m$. We consider two benchmark environments: LunarLander-v2 and Catcher-v0 from GYM \citep{brockman2016openai} and PyGame learning environment \citep{tasfi2016PLE}. We conduct the experiment with 5 independent random trails and report the averaged result with its standard deviation. We refer readers to Appendix \ref{apx_sec: dqn} for more background and other experiment details. Figure \ref{fig: dqn} summarizes the result. For LunarLander-v2, Bootstrap DQN with 2 and 5 heads give similar performance but both converge to a less optimal model compared with the centroid approximation method. Centroid approximation method with 2 and 5 heads performs similarly at convergence but centroid approximation method with 5 heads is able to converge faster than 2-head model and thus has lower regret. For Catcher-v0, adding more heads to the model is able to improve the performance for both methods. The proposed centroid approximation consistently improves over baselines.

\section{Related Work}
\paragraph{Bootstrap} is an classical statistical inference method, which was developed by \citet{efron1992bootstrap} and generalized by, i.e., \citet{mammen1993bootstrap,shao2010dependent,efron2012bayesian} (just to name a few). Bootstrap can be widely applied to various statistical inference problem, such as confidence interval estimation \citep{diciccio1996bootstrap}, model selection \citep{shao1996bootstrap}, high-dimensional inference \citep{chen2018gaussian,el2018can,nie2020bayesian}, off-policy evaluation \citep{hanna2017bootstrapping}, distributed inference \citep{yu2020simultaneous} and inference for ensemble model \citep{kim2020predictive}, etc.

\paragraph{Bayesian Inference} is a different approach to  quantify the model uncertainty. Different from frequentists' method, Bayesian assumes a prior over the model and the uncertainty can be captured by the posterior. 
Bayesian inference have been largely popularized in machine learning, 
largely thanks to the recent development in scalable sampling method \citep{welling2011bayesian,chen2014stochastic,seita2016efficient,wu2020mini}, variational inference \citep{blei2017variational, liu2016stein}, and other approximation methods such as \citet{gal2016dropout,lee2018deep}. 
In comparison, bootstrap has been much less widely used in modern machine learning and deep learning. We believe this is largely attributed to \emph{the lack of similarly efficient computational methods in the small sample size $m$ region}, which is the very problem that we aim to address with our new centroid approximation method.

\paragraph{Uncertainty in Deep Learning}
In additional to the applications considered in this paper, uncertainty in deep learning model can also be applied to problems including calibration \citep{guo2017calibration} and out-of-distribution detection \citep{nguyen2015deep}. The definition of uncertainty of neural network is quite generalized (e.g., \citet{gal2016dropout,ovadia2019can,maddox2019simple,van2020uncertainty}) and can be quite different from the uncertainty that bootstrap inference want to quantify and can be approached with various methods including drop out \citep{gal2016dropout,durasov2020masksembles}, label smoothing \citep{qin2020improving}, designing new modules in the model \citep{kivaranovic2020adaptive}, adversarial training \citep{lakshminarayanan2016simple} and Bayesian modeling \citep{blundell2015weight}, etc. This paper focuses on improving the bootstrap method and thus is orthogonal to those previous works. \citet{pearce2018high, salem2020prediction} also try to refine the ensemble models to improve the quality of prediction interval. Compare with our method, their method can only be applied to prediction interval and does not have theoretical guarantee.

\section{Conclusion}
We propose a centroid approximation method to learn an improved particle distribution that better approximates the target bootstrap distribution, especially in the region with small particle size. Theoretically, when the size of training data is large, our objective function is surrogate to the Wasserstein distance between the particle distribution and target distribution. Thus, compared with standard bootstrap, the proposed centroid approximation method actively optimizes the distance between particle distribution and target distribution. The proposed method is simple and can be flexibly used for applications of bootstrap with negligible extra computational cost.

\nocite{langley00}

\bibliography{ref}

\begin{thebibliography}{63}
\providecommand{\natexlab}[1]{#1}
\providecommand{\url}[1]{\texttt{#1}}
\expandafter\ifx\csname urlstyle\endcsname\relax
  \providecommand{\doi}[1]{doi: #1}\else
  \providecommand{\doi}{doi: \begingroup \urlstyle{rm}\Url}\fi

\bibitem[Austern \& Syrgkanis(2020)Austern and
  Syrgkanis]{austern2020asymptotics}
Austern, M. and Syrgkanis, V.
\newblock Asymptotics of the empirical bootstrap method beyond asymptotic
  normality.
\newblock \emph{arXiv preprint arXiv:2011.11248}, 2020.

\bibitem[Bickel et~al.(2012)Bickel, G{\"o}tze, and van
  Zwet]{bickel2012resampling}
Bickel, P.~J., G{\"o}tze, F., and van Zwet, W.~R.
\newblock Resampling fewer than n observations: gains, losses, and remedies for
  losses.
\newblock In \emph{Selected works of Willem van Zwet}, pp.\  267--297.
  Springer, 2012.

\bibitem[Blei et~al.(2017)Blei, Kucukelbir, and McAuliffe]{blei2017variational}
Blei, D.~M., Kucukelbir, A., and McAuliffe, J.~D.
\newblock Variational inference: A review for statisticians.
\newblock \emph{Journal of the American statistical Association}, 112\penalty0
  (518):\penalty0 859--877, 2017.

\bibitem[Blundell et~al.(2015)Blundell, Cornebise, Kavukcuoglu, and
  Wierstra]{blundell2015weight}
Blundell, C., Cornebise, J., Kavukcuoglu, K., and Wierstra, D.
\newblock Weight uncertainty in neural network.
\newblock In \emph{International Conference on Machine Learning}, pp.\
  1613--1622. PMLR, 2015.

\bibitem[Breiman(1996)]{breiman1996bagging}
Breiman, L.
\newblock Bagging predictors.
\newblock \emph{Machine learning}, 24\penalty0 (2):\penalty0 123--140, 1996.

\bibitem[Brockman et~al.(2016)Brockman, Cheung, Pettersson, Schneider,
  Schulman, Tang, and Zaremba]{brockman2016openai}
Brockman, G., Cheung, V., Pettersson, L., Schneider, J., Schulman, J., Tang,
  J., and Zaremba, W.
\newblock Openai gym.
\newblock \emph{arXiv preprint arXiv:1606.01540}, 2016.

\bibitem[Campbell \& Beronov(2019)Campbell and Beronov]{campbell2019sparse}
Campbell, T. and Beronov, B.
\newblock Sparse variational inference: Bayesian coresets from scratch.
\newblock \emph{Advances in Neural Information Processing Systems},
  32:\penalty0 11461--11472, 2019.

\bibitem[Canas \& Rosasco(2012)Canas and Rosasco]{NIPS2012_c54e7837}
Canas, G. and Rosasco, L.
\newblock Learning probability measures with respect to optimal transport
  metrics.
\newblock In Pereira, F., Burges, C. J.~C., Bottou, L., and Weinberger, K.~Q.
  (eds.), \emph{Advances in Neural Information Processing Systems}, volume~25.
  Curran Associates, Inc., 2012.

\bibitem[Chatterjee et~al.(2005)Chatterjee, Bose,
  et~al.]{chatterjee2005generalized}
Chatterjee, S., Bose, A., et~al.
\newblock Generalized bootstrap for estimating equations.
\newblock \emph{The Annals of Statistics}, 33\penalty0 (1):\penalty0 414--436,
  2005.

\bibitem[Chen et~al.(2014)Chen, Fox, and Guestrin]{chen2014stochastic}
Chen, T., Fox, E., and Guestrin, C.
\newblock Stochastic gradient hamiltonian monte carlo.
\newblock In \emph{International conference on machine learning}, pp.\
  1683--1691. PMLR, 2014.

\bibitem[Chen et~al.(2018{\natexlab{a}})Chen, Mackey, Gorham, Briol, and
  Oates]{chen2018stein}
Chen, W.~Y., Mackey, L., Gorham, J., Briol, F.-X., and Oates, C.
\newblock Stein points.
\newblock In \emph{International Conference on Machine Learning}, pp.\
  844--853. PMLR, 2018{\natexlab{a}}.

\bibitem[Chen et~al.(2018{\natexlab{b}})]{chen2018gaussian}
Chen, X. et~al.
\newblock Gaussian and bootstrap approximations for high-dimensional
  u-statistics and their applications.
\newblock \emph{The Annals of Statistics}, 46\penalty0 (2):\penalty0 642--678,
  2018{\natexlab{b}}.

\bibitem[Chen et~al.(2012)Chen, Welling, and Smola]{chen2012super}
Chen, Y., Welling, M., and Smola, A.
\newblock Super-samples from kernel herding.
\newblock \emph{arXiv preprint arXiv:1203.3472}, 2012.

\bibitem[Cheng et~al.(2010)Cheng, Huang, et~al.]{cheng2010bootstrap}
Cheng, G., Huang, J.~Z., et~al.
\newblock Bootstrap consistency for general semiparametric m-estimation.
\newblock \emph{Annals of Statistics}, 38\penalty0 (5):\penalty0 2884--2915,
  2010.

\bibitem[Claici et~al.(2018)Claici, Genevay, and
  Solomon]{claici2018wasserstein}
Claici, S., Genevay, A., and Solomon, J.
\newblock Wasserstein measure coresets.
\newblock \emph{arXiv preprint arXiv:1805.07412}, 2018.

\bibitem[DiCiccio et~al.(1996)DiCiccio, Efron, et~al.]{diciccio1996bootstrap}
DiCiccio, T.~J., Efron, B., et~al.
\newblock Bootstrap confidence intervals.
\newblock \emph{Statistical science}, 11\penalty0 (3):\penalty0 189--228, 1996.

\bibitem[Durasov et~al.(2020)Durasov, Bagautdinov, Baque, and
  Fua]{durasov2020masksembles}
Durasov, N., Bagautdinov, T., Baque, P., and Fua, P.
\newblock Masksembles for uncertainty estimation.
\newblock \emph{arXiv preprint arXiv:2012.08334}, 2020.

\bibitem[Efron(1992)]{efron1992bootstrap}
Efron, B.
\newblock Bootstrap methods: another look at the jackknife.
\newblock In \emph{Breakthroughs in statistics}, pp.\  569--593. Springer,
  1992.

\bibitem[Efron(2012)]{efron2012bayesian}
Efron, B.
\newblock Bayesian inference and the parametric bootstrap.
\newblock \emph{The annals of applied statistics}, 6\penalty0 (4):\penalty0
  1971, 2012.

\bibitem[El~Karoui \& Purdom(2018)El~Karoui and Purdom]{el2018can}
El~Karoui, N. and Purdom, E.
\newblock Can we trust the bootstrap in high-dimensions? the case of linear
  models.
\newblock \emph{The Journal of Machine Learning Research}, 19\penalty0
  (1):\penalty0 170--235, 2018.

\bibitem[Gal \& Ghahramani(2016)Gal and Ghahramani]{gal2016dropout}
Gal, Y. and Ghahramani, Z.
\newblock Dropout as a bayesian approximation: Representing model uncertainty
  in deep learning.
\newblock In \emph{international conference on machine learning}, pp.\
  1050--1059. PMLR, 2016.

\bibitem[Gong \& Ye(2021)Gong and Ye]{gong2021argmax}
Gong, C. and Ye, M.
\newblock Argmax centroids: with applications to multi-domain learning.
\newblock \emph{NeurIPS 2021}, 2021.

\bibitem[Graves(2011)]{graves2011practical}
Graves, A.
\newblock Practical variational inference for neural networks.
\newblock In \emph{Advances in neural information processing systems}, pp.\
  2348--2356. Citeseer, 2011.

\bibitem[Guo et~al.(2017)Guo, Pleiss, Sun, and Weinberger]{guo2017calibration}
Guo, C., Pleiss, G., Sun, Y., and Weinberger, K.~Q.
\newblock On calibration of modern neural networks.
\newblock In \emph{International Conference on Machine Learning}, pp.\
  1321--1330. PMLR, 2017.

\bibitem[Hall et~al.(1988)]{hall1988rate}
Hall, P. et~al.
\newblock Rate of convergence in bootstrap approximations.
\newblock \emph{The Annals of Probability}, 16\penalty0 (4):\penalty0
  1665--1684, 1988.

\bibitem[Hanna et~al.(2017)Hanna, Stone, and Niekum]{hanna2017bootstrapping}
Hanna, J., Stone, P., and Niekum, S.
\newblock Bootstrapping with models: Confidence intervals for off-policy
  evaluation.
\newblock In \emph{Proceedings of the AAAI Conference on Artificial
  Intelligence}, volume~31, 2017.

\bibitem[Hao et~al.(2019)Hao, Abbasi~Yadkori, Wen, and
  Cheng]{hao2019bootstrapping}
Hao, B., Abbasi~Yadkori, Y., Wen, Z., and Cheng, G.
\newblock Bootstrapping upper confidence bound.
\newblock In Wallach, H., Larochelle, H., Beygelzimer, A., d\textquotesingle
  Alch\'{e}-Buc, F., Fox, E., and Garnett, R. (eds.), \emph{Advances in Neural
  Information Processing Systems}, volume~32. Curran Associates, Inc., 2019.

\bibitem[Hron et~al.(2017)Hron, Matthews, and Ghahramani]{hron2017variational}
Hron, J., Matthews, A. G. d.~G., and Ghahramani, Z.
\newblock Variational gaussian dropout is not bayesian.
\newblock \emph{arXiv preprint arXiv:1711.02989}, 2017.

\bibitem[Huang et~al.(2017)Huang, Li, Pleiss, Liu, Hopcroft, and
  Weinberger]{huang2017snapshot}
Huang, G., Li, Y., Pleiss, G., Liu, Z., Hopcroft, J.~E., and Weinberger, K.~Q.
\newblock Snapshot ensembles: Train 1, get m for free.
\newblock \emph{International Conference on Learning Representations}, 2017.

\bibitem[Kim et~al.(2020)Kim, Xu, and Foygel~Barber]{kim2020predictive}
Kim, B., Xu, C., and Foygel~Barber, R.
\newblock Predictive inference is free with the jackknife+-after-bootstrap.
\newblock \emph{Advances in Neural Information Processing Systems}, 33, 2020.

\bibitem[Kivaranovic et~al.(2020)Kivaranovic, Johnson, and
  Leeb]{kivaranovic2020adaptive}
Kivaranovic, D., Johnson, K.~D., and Leeb, H.
\newblock Adaptive, distribution-free prediction intervals for deep networks.
\newblock In \emph{International Conference on Artificial Intelligence and
  Statistics}, pp.\  4346--4356. PMLR, 2020.

\bibitem[Kleiner et~al.(2014)Kleiner, Talwalkar, Sarkar, and
  Jordan]{kleiner2014scalable}
Kleiner, A., Talwalkar, A., Sarkar, P., and Jordan, M.~I.
\newblock A scalable bootstrap for massive data.
\newblock \emph{Journal of the Royal Statistical Society: Series B: Statistical
  Methodology}, pp.\  795--816, 2014.

\bibitem[Lakshminarayanan et~al.(2017)Lakshminarayanan, Pritzel, and
  Blundell]{lakshminarayanan2016simple}
Lakshminarayanan, B., Pritzel, A., and Blundell, C.
\newblock Simple and scalable predictive uncertainty estimation using deep
  ensembles.
\newblock \emph{Advances in neural information processing systems}, 30, 2017.

\bibitem[Lee et~al.(2018)Lee, Sohl-dickstein, Pennington, Novak, Schoenholz,
  and Bahri]{lee2018deep}
Lee, J., Sohl-dickstein, J., Pennington, J., Novak, R., Schoenholz, S., and
  Bahri, Y.
\newblock Deep neural networks as gaussian processes.
\newblock In \emph{International Conference on Learning Representations}, 2018.

\bibitem[Liu \& Wang(2016)Liu and Wang]{liu2016stein}
Liu, Q. and Wang, D.
\newblock Stein variational gradient descent: A general purpose bayesian
  inference algorithm.
\newblock \emph{Advances in Neural Information Processing Systems}, 29, 2016.

\bibitem[Maddox et~al.(2019)Maddox, Izmailov, Garipov, Vetrov, and
  Wilson]{maddox2019simple}
Maddox, W.~J., Izmailov, P., Garipov, T., Vetrov, D.~P., and Wilson, A.~G.
\newblock A simple baseline for bayesian uncertainty in deep learning.
\newblock \emph{Advances in Neural Information Processing Systems},
  32:\penalty0 13153--13164, 2019.

\bibitem[Mammen(1993)]{mammen1993bootstrap}
Mammen, E.
\newblock Bootstrap and wild bootstrap for high dimensional linear models.
\newblock \emph{The annals of statistics}, pp.\  255--285, 1993.

\bibitem[May et~al.(2012)May, Korda, Lee, and Leslie]{may2012optimistic}
May, B.~C., Korda, N., Lee, A., and Leslie, D.~S.
\newblock Optimistic bayesian sampling in contextual-bandit problems.
\newblock \emph{Journal of Machine Learning Research}, 13:\penalty0 2069--2106,
  2012.

\bibitem[Nguyen et~al.(2015)Nguyen, Yosinski, and Clune]{nguyen2015deep}
Nguyen, A., Yosinski, J., and Clune, J.
\newblock Deep neural networks are easily fooled: High confidence predictions
  for unrecognizable images.
\newblock In \emph{Proceedings of the IEEE conference on computer vision and
  pattern recognition}, pp.\  427--436, 2015.

\bibitem[Nie \& Ro{\v{c}}kov{\'a}(2020)Nie and
  Ro{\v{c}}kov{\'a}]{nie2020bayesian}
Nie, L. and Ro{\v{c}}kov{\'a}, V.
\newblock Bayesian bootstrap spike-and-slab lasso.
\newblock \emph{arXiv preprint arXiv:2011.14279}, 2020.

\bibitem[Osband et~al.(2016)Osband, Blundell, Pritzel, and
  Van~Roy]{osband2016deep}
Osband, I., Blundell, C., Pritzel, A., and Van~Roy, B.
\newblock Deep exploration via bootstrapped dqn.
\newblock \emph{Advances in neural information processing systems},
  29:\penalty0 4026--4034, 2016.

\bibitem[Ovadia et~al.(2019)Ovadia, Fertig, Ren, Nado, Sculley, Nowozin,
  Dillon, Lakshminarayanan, and Snoek]{ovadia2019can}
Ovadia, Y., Fertig, E., Ren, J., Nado, Z., Sculley, D., Nowozin, S., Dillon,
  J., Lakshminarayanan, B., and Snoek, J.
\newblock Can you trust your model's uncertainty? evaluating predictive
  uncertainty under dataset shift.
\newblock \emph{Advances in Neural Information Processing Systems},
  32:\penalty0 13991--14002, 2019.

\bibitem[Pearce et~al.(2018)Pearce, Brintrup, Zaki, and Neely]{pearce2018high}
Pearce, T., Brintrup, A., Zaki, M., and Neely, A.
\newblock High-quality prediction intervals for deep learning: A
  distribution-free, ensembled approach.
\newblock In \emph{International Conference on Machine Learning}, pp.\
  4075--4084. PMLR, 2018.

\bibitem[Qin et~al.(2020)Qin, Wang, Beutel, and Chi]{qin2020improving}
Qin, Y., Wang, X., Beutel, A., and Chi, E.~H.
\newblock Improving uncertainty estimates through the relationship with
  adversarial robustness.
\newblock \emph{arXiv preprint arXiv:2006.16375}, 2020.

\bibitem[Riquelme et~al.(2018)Riquelme, Tucker, and Snoek]{riquelme2018deep}
Riquelme, C., Tucker, G., and Snoek, J.
\newblock Deep bayesian bandits showdown: An empirical comparison of bayesian
  deep networks for thompson sampling.
\newblock In \emph{International Conference on Learning Representations}, 2018.

\bibitem[Rubin(1981)]{rubin1981bayesian}
Rubin, D.~B.
\newblock The bayesian bootstrap.
\newblock \emph{The annals of statistics}, pp.\  130--134, 1981.

\bibitem[Salem et~al.(2020)Salem, Langseth, and
  Ramampiaro]{salem2020prediction}
Salem, T.~S., Langseth, H., and Ramampiaro, H.
\newblock Prediction intervals: Split normal mixture from quality-driven deep
  ensembles.
\newblock In \emph{Conference on Uncertainty in Artificial Intelligence}, pp.\
  1179--1187. PMLR, 2020.

\bibitem[Seita et~al.(2018)Seita, Pan, Chen, and Canny]{seita2016efficient}
Seita, D., Pan, X., Chen, H., and Canny, J.
\newblock An efficient minibatch acceptance test for metropolis-hastings.
\newblock In \emph{Proceedings of the 27th International Joint Conference on
  Artificial Intelligence}, pp.\  5359--5363, 2018.

\bibitem[Shao(1996)]{shao1996bootstrap}
Shao, J.
\newblock Bootstrap model selection.
\newblock \emph{Journal of the American statistical Association}, 91\penalty0
  (434):\penalty0 655--665, 1996.

\bibitem[Shao(2010)]{shao2010dependent}
Shao, X.
\newblock The dependent wild bootstrap.
\newblock \emph{Journal of the American Statistical Association}, 105\penalty0
  (489):\penalty0 218--235, 2010.

\bibitem[Simonyan \& Zisserman(2014)Simonyan and Zisserman]{simonyan2014very}
Simonyan, K. and Zisserman, A.
\newblock Very deep convolutional networks for large-scale image recognition.
\newblock \emph{arXiv preprint arXiv:1409.1556}, 2014.

\bibitem[Srivastava et~al.(2014)Srivastava, Hinton, Krizhevsky, Sutskever, and
  Salakhutdinov]{srivastava2014dropout}
Srivastava, N., Hinton, G., Krizhevsky, A., Sutskever, I., and Salakhutdinov,
  R.
\newblock Dropout: a simple way to prevent neural networks from overfitting.
\newblock \emph{The journal of machine learning research}, 15\penalty0
  (1):\penalty0 1929--1958, 2014.

\bibitem[Tasfi(2016)]{tasfi2016PLE}
Tasfi, N.
\newblock Pygame learning environment.
\newblock \url{https://github.com/ntasfi/PyGame-Learning-Environment}, 2016.

\bibitem[Thompson(1933)]{thompson1933likelihood}
Thompson, W.~R.
\newblock On the likelihood that one unknown probability exceeds another in
  view of the evidence of two samples.
\newblock \emph{Biometrika}, 25\penalty0 (3/4):\penalty0 285--294, 1933.

\bibitem[Van~Amersfoort et~al.(2020)Van~Amersfoort, Smith, Teh, and
  Gal]{van2020uncertainty}
Van~Amersfoort, J., Smith, L., Teh, Y.~W., and Gal, Y.
\newblock Uncertainty estimation using a single deep deterministic neural
  network.
\newblock In \emph{International Conference on Machine Learning}, pp.\
  9690--9700. PMLR, 2020.

\bibitem[Van~Hasselt et~al.(2016)Van~Hasselt, Guez, and Silver]{van2016deep}
Van~Hasselt, H., Guez, A., and Silver, D.
\newblock Deep reinforcement learning with double q-learning.
\newblock In \emph{Proceedings of the AAAI Conference on Artificial
  Intelligence}, volume~30, 2016.

\bibitem[Vyas et~al.(2018)Vyas, Jammalamadaka, Zhu, Das, Kaul, and
  Willke]{vyas2018out}
Vyas, A., Jammalamadaka, N., Zhu, X., Das, D., Kaul, B., and Willke, T.~L.
\newblock Out-of-distribution detection using an ensemble of self supervised
  leave-out classifiers.
\newblock In \emph{Proceedings of the European Conference on Computer Vision
  (ECCV)}, pp.\  550--564, 2018.

\bibitem[Wasserman(2013)]{wasserman2013all}
Wasserman, L.
\newblock \emph{All of statistics: a concise course in statistical inference}.
\newblock Springer Science \& Business Media, 2013.

\bibitem[Weed et~al.(2019)Weed, Bach, et~al.]{weed2019sharp}
Weed, J., Bach, F., et~al.
\newblock Sharp asymptotic and finite-sample rates of convergence of empirical
  measures in wasserstein distance.
\newblock \emph{Bernoulli}, 25\penalty0 (4A):\penalty0 2620--2648, 2019.

\bibitem[Welling \& Teh(2011)Welling and Teh]{welling2011bayesian}
Welling, M. and Teh, Y.~W.
\newblock Bayesian learning via stochastic gradient langevin dynamics.
\newblock In \emph{Proceedings of the 28th international conference on machine
  learning (ICML-11)}, pp.\  681--688. Citeseer, 2011.

\bibitem[Wu et~al.(2020)Wu, Rachel~Wang, and Wong]{wu2020mini}
Wu, T.-Y., Rachel~Wang, Y., and Wong, W.~H.
\newblock Mini-batch metropolis--hastings with reversible sgld proposal.
\newblock \emph{Journal of the American Statistical Association}, pp.\  1--9,
  2020.

\bibitem[Wyatt(1998)]{wyatt1998exploration}
Wyatt, J.
\newblock Exploration and inference in learning from reinforcement.
\newblock 1998.

\bibitem[Yu et~al.(2020)Yu, Chao, and Cheng]{yu2020simultaneous}
Yu, Y., Chao, S.-K., and Cheng, G.
\newblock Simultaneous inference for massive data: Distributed bootstrap.
\newblock In \emph{International Conference on Machine Learning}, pp.\
  10892--10901. PMLR, 2020.

\end{thebibliography}
\bibliographystyle{icml2022}

\newpage
\appendix
\onecolumn
\section{Algorithm Box} \label{apx_sec: alpo}
In practical implementation, we do not need to update $\w_h$ every iteration and can also replace the full-batch gradient by stochastic gradient. Specifically, notice that $\hat{g(\th_j^*)}$ defined in (\ref{equ: g_hat}) can be alternative represented as 
\begin{align} \label{equ: g_hat_2}
    \hat{g}(\th_{j}^{*}(t))=\frac{\sum_{h=1}^{M}\sum_{i=1}^{n}\left[\mathbb{I}\{j\in u_{\w_{h}}(t)\}\right]w_{h,i}\nabla_{\th}\ell(x_{i},f_{\th^*_j(t)})/n}{\sum_{h=1}^{M}\left[\mathbb{I}\{j\in u_{\w_{h}}\}\right]}=\frac{1}{n}\sum_{i=1}^{n}q_{i,j}\nabla_{\th}\ell(x_{i},f_{\th_j^*(t)}),
\end{align}
where $q_{i,j}$ is defined by 
\begin{align} \label{equ: qij}
    q_{ij}:=\frac{\sum_{h=1}^{M}\left[\mathbb{I}\{j\in u_{\w_{h}}(t)\}\right]w_{h,i}}{\sum_{h=1}^{M}\left[\mathbb{I}\{j\in u_{\w_{h}}(t)\}\right]}.
\end{align}
This allows us to use a stochastic gradient version of gradient
\begin{align} \label{equ: sgd}
    \hat{g}_{\text{sgd}}(\th_{j}^{*})=\frac{1}{|B|}\sum_{i\in[B]}q_{i,j}\nabla_{\th}\ell(x_{i},f_{\th_{j}^{*}}),
\end{align}
where $B$ is the set of mini-batch data. The detailed algorithm is summarized in Algorithm \ref{algo}

\begin{algorithm}[t]
\begin{algorithmic}[1]
\STATE{Initialize $\th^*_j(0)$, $j\in[m]$ by i.i.d. sampling from $\rho_\pi$ or other distribution such as Gaussian.}
\FOR{$t \in \text{iterations}$}
    \STATE{// Update $\w_h$ only every a few iterations.}
    \IF{$t$ \text{mod freq} $==0$}
        \STATE{$\forall j\in[m]$, calculate $\LL(\th^*_j(t))$ defined in (\ref{equ: datawise loss})}
        \STATE{Sample $\{\w_h\}_{h=1}^M$, i.i.d. from $\pi$.}
        \STATE{Calculate $\LL_{\w_h}(\th^*_j(t))=\w_h^T\LL(\th_j^*(t))$, 
        $\forall h \in [M]$ and $j \in [m]$, }
        \STATE{$\forall h \in [M]$, calculate $u_{\w_h}(t)$ defined in (\ref{equ: uwh}).}
    \ELSE
        \STATE{$u_{\w_h}(t)$ = $u_{\w_h}(t-1)$}
    \ENDIF
    \STATE{$\forall j \in [m]$, update $\th^*_j(t)$ by (\ref{update: practical}). {(May use mini-batch gradient defined in (\ref{equ: sgd}))}. }
\ENDFOR
\end{algorithmic}\caption{Practical implementation of centroid approximation with less frequent updating of $\w_h$ and stochastic gradient enabled.} \label{algo}
\end{algorithm}

\section{Additional Experiment details}
\subsection{Bootstrap Confidence Interval} \label{apx_sec: ci}

Given a model $f_{\th}$ parameterized by $\th$ and a training set with $n$ data points i.i.d. sampled from population, our goal is to construct confidence interval for $\th$. Let $\tilde\rho_\pi$ be an empirical distribution approximating $\rho_\pi$, which could be obtained by i.i.d. sampling, or by our centroid method. Denote by $Q[\alpha, \tilde\rho_\pi]$ the $\alpha$-quantile function of $\tilde \rho_\pi$  with some $\alpha\in[0,1]$. We consider the following three ways to construct (two-sided) bootstrap confidence interval of $\th$ with confidence level $\alpha$: the Normal interval, the percentile interval and the pivotal interval which are defined below. 

\paragraph{Methods to construct confidence interval}
The methods we used to construct confidence interval are
-- The Normal interval:
\[
[\hat{\th}-z((1+\alpha)/2)\hat{\text{se}}_{\text{boot}},\hat{\th}+z((1+\alpha)/2)\hat{\text{se}}_{\text{boot}}],
\]
where $z(\cdot)$ is the inverse cumulative distribution function of standard Normal distribution.
And $\hat{\text{se}}_{\text{boot}}$ is the standard deviation estimated from $\tilde \rho_\pi$.

-- The percentile intervals:
\[
[Q[(1-\alpha)/2,\tilde{\rho}_{\pi}],Q[(1+\alpha)/2,\tilde{\rho}_{\pi}]].
\]
-- The pivotal interval:
\[
[2\hat{\th}-Q[(1+\alpha)/2,\tilde{\rho}_{\pi}],2\hat{\th}-Q[(1-\alpha)/2,\tilde{\rho}_{\pi}]].
\]

We consider the following simple linear regression: $x  \sim\N(\boldsymbol{0},\boldsymbol{I})$,
$y\mid x \sim\mathbb{\N}(\theta^\top  x,\boldsymbol{I})$,
where the features $x\in\mathbb{R}^4$ and we set the true parameter to be $\theta_0 =[1,-1,1,-1]$. We consider $n=50$ and the number of particles $m=20,50,100,200$. We compare the coverage probability and the confidence level $\alpha$ to measure the quality:

\paragraph{Measuring the quality of confidence interval}
With a large number $N$ of independently generated training data (we use $N=1000$), we are able to obtain the corresponding confidence intervals $\{\text{CI}(\alpha)_s\}_{s=1}^{N} $and thus obtain the probability that the true parameter falls into the confidence intervals, which is the estimated coverage probability
\vspace{-0.2cm}
\[
\hat{\alpha} = \frac{1}{N}\sum_{s=1}^{N}\mathbb{I}\{\th_0 \in \text{CI}(\alpha)_s \} .
\] A good confidence interval should have $\hat{\alpha}$ close to $\alpha$. Thus we measure the performance by calculating the difference $|\alpha-\hat{\alpha}|$.

As $\hat{\th}_{\w}$ is the least square estimator of the bootstrapped
dataset, it has analytic solution and thus can be obtained via some
matrix multiplications. $\th_{\w}^{*}$ is initialized using $\hat{\th}_{\w}$
and then updated for 2000 steps. For this experiment, we find that
adding the threshold $\gamma$ does not gives further improvement
for this experiment and thus we simply set $\gamma=0$ and use $M=1$. We approximate
the true bootstrap distribution by sampling 10000 i.i.d. samples.

\paragraph{More experiment result}
Table \ref{tbl: ci_all} all the result we have varying $\alpha=0.8,0.9,0.95$, $m=20,50,100,200$ and three different approaches for constructing confidence interval. As we can see, centroid approximation gives the best performance in most cases compared with the other three baselines.

\begin{table}
\centering{}%
\scalebox{0.9}{
\begin{tabular}{c|c|c|cccc}
\toprule
\multicolumn{1}{c}{} & \multicolumn{2}{c|}{Num Particle} & 20 & 50 & 100 & 200\tabularnewline
\hline 
\multirow{12}{*}{$\alpha=0.8$} & \multirow{4}{*}{Normal} & Bootstrap & $0.033\pm0.013$ & $0.028\pm0.013$ & $0.026\pm0.013$ & $0.031\pm0.013$\tabularnewline
 &  & Bayesian & $0.084\pm0.014$ & $0.076\pm0.014$ & $0.082\pm0..014$ & $0.086\pm0.014$\tabularnewline
 &  & Residual & $\textbf{0.033}\pm\textbf{0.013}$ & $0.037\pm0.013$ & $0.029\pm0.013$ & $\textbf{0.024}\pm\textbf{0.013}$\tabularnewline
 &  & Centroid & $0.036\pm0.013$ & $\textbf{0.003}\pm\textbf{0.012}$ & $\textbf{0.017}\pm\textbf{0.013}$ & $0.030\pm0.013$\tabularnewline
\cline{2-7} \cline{3-7} \cline{4-7} \cline{5-7} \cline{6-7} \cline{7-7} 
 & \multirow{4}{*}{Percentile} & Bootstrap & $0.096\pm0.014$ & $0.050\pm0.014$ & $0.044\pm0.013$ & $0.024\pm0.013$\tabularnewline
 &  & Bayesian & $0.114\pm0.015$ & $0.079\pm0.014$ & $0.074\pm0.014$ & $0.071\pm0.014$\tabularnewline
 &  & Residual & $0.079\pm0.014$ & $0.032\pm0.013$ & $\textbf{0.017}\pm\textbf{0.013}$ & $\textbf{0.010}\pm\textbf{0.013}$\tabularnewline
 &  & Centroid & $\textbf{0.066}\pm\textbf{0.014}$ & $\textbf{0.008}\pm\textbf{0.013}$ & $0.019\pm0.013$ & $0.020\pm0.013$\tabularnewline
\cline{2-7} \cline{3-7} \cline{4-7} \cline{5-7} \cline{6-7} \cline{7-7} 
 & \multirow{4}{*}{Pivotal} & Bootstrap & $0.101\pm0.015$ & $0.053\pm0.014$ & $0.045\pm0.014$ & $0.033\pm0.013$\tabularnewline
 &  & Bayesian & $0.158\pm0.015$ & $0.110\pm0.110$ & $0.088\pm0.014$ & $0.078\pm0.014$\tabularnewline
 &  & Residual & $0.087\pm0.014$ & $0.044\pm0.013$ & $0.030\pm0.013$ & $\textbf{0.023}\pm\textbf{0.013}$\tabularnewline
 &  & Centroid & $\textbf{0.026}\pm\textbf{0.013}$ & $\textbf{0.030}\pm\textbf{0.012}$ & $\textbf{0.018}\pm\textbf{0.013}$ & $0.030\pm0.013$\tabularnewline
\hline 
\multirow{12}{*}{$\alpha=0.9$} & \multirow{4}{*}{Normal} & Bootstrap & $0.029\pm0.010$ & $0.031\pm0.011$ & $0.021\pm0.010$ & $0.017\pm0.010$\tabularnewline
 &  & Bayesian & $0.076\pm0.012$ & $0.054\pm0.011$ & $0.048\pm0.011$ & $0.045\pm0.011$\tabularnewline
 &  & Residual & $0.043\pm0.011$ & $0.023\pm0.010$ & $0.025\pm0.010$ & $0.020\pm0.010$\tabularnewline
 &  & Centroid & $\textbf{0.027}\pm\textbf{0.010}$ & $\textbf{0.001}\pm\textbf{0.009}$ & $\textbf{0.012}\pm\textbf{0.010}$ & $\textbf{0.016}\pm\textbf{0.010}$\tabularnewline
\cline{2-7} \cline{3-7} \cline{4-7} \cline{5-7} \cline{6-7} \cline{7-7} 
 & \multirow{4}{*}{Percentile} & Bootstrap & $0.101\pm0.013$ & $0.036\pm0.011$ & $0.021\pm0.010$ & $\textbf{0.014}\pm\textbf{0.010}$\tabularnewline
 &  & Bayesian & $0.129\pm0.013$ & $0.077\pm0.012$ & $0.059\pm0.012$ & $0.054\pm0.011$\tabularnewline
 &  & Residual & $0.098\pm0.013$ & $0.030\pm0.011$ & $0.033\pm0.011$ & $0.025\pm0.010$\tabularnewline
 &  & Centroid & $\textbf{0.081}\pm\textbf{0.012}$ & $\textbf{0.021}\pm\textbf{0.010}$ & $\textbf{0.020}\pm\textbf{0.010}$ & $0.015\pm0.010$\tabularnewline
\cline{2-7} \cline{3-7} \cline{4-7} \cline{5-7} \cline{6-7} \cline{7-7} 
 & \multirow{4}{*}{Pivotal} & Bootstrap & $0.106\pm0.013$ & $0.045\pm0.011$ & $0.025\pm0.010$ & $0.023\pm0.010$\tabularnewline
 &  & Bayesian & $0.149\pm0.014$ & $0.093\pm0.012$ & $0.073\pm0.012$ & $0.056\pm0.011$\tabularnewline
 &  & Residual & $0.100\pm0.013$ & $0.044\pm0.011$ & $0.030\pm0.011$ & $0.023\pm0.010$\tabularnewline
 &  & Centroid & $\textbf{0.046}\pm\textbf{0.011}$ & $\textbf{0.013}\pm\textbf{0.009}$ & $\textbf{0.011}\pm\textbf{0.010}$ & $\textbf{0.020}\pm\textbf{0.010}$\tabularnewline
\hline 
\multirow{12}{*}{$\alpha=0.95$} & \multirow{4}{*}{Normal} & Bootstrap & $\textbf{0.018}\pm\textbf{0.008}$ & $0.014\pm0.008$ & $0.012\pm0.008$ & $0.006\pm0.007$\tabularnewline
 &  & Bayesian & $0.053\pm0.010$ & $0.038\pm0.009$ & $0.031\pm0.009$ & $0.037\pm0.009$\tabularnewline
 &  & Residual & $0.036\pm0.009$ & $0.019\pm0.008$ & $0.011\pm0.008$ & $0.008\pm0.007$\tabularnewline
 &  & Centroid & $\textbf{0.018}\pm\textbf{0.008}$ & $\textbf{0.005}\pm\textbf{0.006}$ & $\textbf{0.009}\pm\textbf{0.007}$ & $\textbf{0.005}\pm\textbf{0.007}$\tabularnewline
\cline{2-7} \cline{3-7} \cline{4-7} \cline{5-7} \cline{6-7} \cline{7-7} 
 & \multirow{4}{*}{Percentile} & Bootstrap & $0.081\pm0.010$ & $0.047\pm0.009$ & $0.030\pm0.008$ & $0.017\pm0.008$\tabularnewline
 &  & Bayesian & $0.126\pm0.012$ & $0.072\pm0.010$ & $0.056\pm0.010$ & $0.042\pm0.009$\tabularnewline
 &  & Residual & $0.100\pm0.011$ & $0.040\pm0.009$ & $0.037\pm0.009$ & $0.021\pm0.008$\tabularnewline
 &  & Centroid & $\textbf{0.077}\pm\textbf{0.010}$ & $\textbf{0.029}\pm\textbf{0.008}$ & $\textbf{0.020}\pm\textbf{0.008}$ & $\textbf{0.016}\pm\textbf{0.008}$\tabularnewline
\cline{2-7} \cline{3-7} \cline{4-7} \cline{5-7} \cline{6-7} \cline{7-7} 
 & \multirow{4}{*}{Pivotal} & Bootstrap & $0.089\pm0.011$ & $0.043\pm0.009$ & $0.027\pm0.008$ & $0.015\pm0.008$\tabularnewline
 &  & Bayesian & $0.127\pm0.012$ & $0.091\pm0.011$ & $0.064\pm0.010$ & $0.056\pm0.010$\tabularnewline
 &  & Residual & $0.085\pm0.011$ & $0.051\pm0.009$ & $0.036\pm0.009$ & $0.029\pm0.008$\tabularnewline
 &  & Centroid & $\textbf{0.046}\pm\textbf{0.009}$ & $\textbf{0.002}\pm\textbf{0.007}$ & $\textbf{0.014}\pm\textbf{0.008}$ & $\textbf{0.009}\pm\textbf{0.007}$\tabularnewline
\bottomrule
\end{tabular}
}\caption{Complete result on comparing centroid approximation with various bootstrap methods. The bold number shows the best approach.} \label{tbl: ci_all}
\end{table}

\subsection{Centroid Approximation for Bootstrap Method in Contextual Bandit} \label{apx_sec: cb}

\subsubsection{More background}
Contextual bandit is a classic task in sequential decision making
problem in which at time $t=1,...,n$, a new context $x_{t}$ arrives
and is observed by an algorithm. Based on its internal model, the
algorithm selects an actions $a_{t}$ and receives a reward $r_{t}(x_{t},a_{t})$
related to the context and action. During this process, the algorithm
may update its internal model with the newly received data. At the
end of this process, the cumulative reward of the algorithm is calculated
by $r=\sum_{t=1}^{n}r_{t}$ and the goal for the algorithm is to improve
the cumulative reward $r$. The exploration-exploitation dilemma is
a fundamental aspect in sequential decision making problem such as
contextual bandit: the algorithm needs to trade-off between the best
expected action returned by the internal model at the moment (i.e.,
exploitation) with potentially sub-optimal exploratory actions. Thompson
sampling \citep{thompson1933likelihood,wyatt1998exploration,may2012optimistic} is an elegant and effective approach to tackle
the exploration-exploitation dilemma using the model uncertainty,
which can be approached with various methods including Bayesian posterior \citep{graves2011practical,welling2011bayesian}, dropout uncertainty \citep{srivastava2014dropout,hron2017variational} and Bootstrap \citep{osband2016deep,hao2019bootstrapping}. The ability to accurately assess the uncertainty is a key to improve the cumulative reward. Bootstrap method for contextual bandit maintains $m$ bootstrap models trained with different bootstrapped training set. When conducting an action, the algorithm uniformly samples a model and then selects the best action returned by the sampled model. 

\subsubsection{More experiment setup details}
{\color{black}
We set all the experimental setting including data preprocessing, network architecture and training pipeline exactly the same as the one used in \citet{riquelme2018deep} and adopt its open source code repository.

\textbf{Network architecture}
Following \citet{riquelme2018deep}, we consider a fully connected feed forward network with two hidden layers with 50 hidden units and ReLU activations. The input and output dimensions
depends on the dimension of context and number of possible actions.

\begin{algorithm}[t]
\begin{algorithmic}[1]
\STATE{Obtain a randomly initialized $\th^*_j(0)$, $j\in[m]$.}
\STATE{Initialize a common replay buffer $R_c = \emptyset$ recording all the observed contexts.}
\STATE{For each model, initialize its own replay buffer $R_j = \emptyset$ that is used for training.}
\FOR{$t \in \text{number of total steps}$}
    \STATE{Obtain the $t$-th context $x_t$.}
    \STATE{Sampling one model based on probability $\{v_j^*(t)\}_{j=1}^m$ to make action $a_t$ and get reward $r_t$.}
    \STATE{Update the common replay buffer by $R_{c}\leftarrow R_{c}\cup\{(x_t, a_t, r_t)\}$}
    \STATE{// Update $\w_h$ and $R_j$ and model every a few iterations.}
    \IF{$t$ \text{mod freq} $== 0$}
        \STATE{$\forall j\in[m]$, calculate $\LL(\th^*_j(t))$ defined in (\ref{equ: datawise loss}) for all the contexts in $R_c$. // \scriptsize{$\LL(\th^*_j(t)) \in \R^{|R_c|}$.}}
        \STATE{Generate $M$ sets of random weights $\{\w_h\}_{h=1}^M$ of contexts in $R_c$ from $\pi$.}
        \STATE{$\forall h \in [M]$ and $j \in [m]$, calculate $\LL_{\w_h}(\th^*_j(t))=\w_h^T\LL(\th_j^*(t))$.}
        \STATE{$\forall h \in [M]$, calculate $u_{\w_h}(t)$ defined in (\ref{equ: uwh}) for each $h$.}
        \STATE{$\forall i \in [|R_c|]$ and $j \in [m]$, calculate $q_{i,j}$ by (\ref{equ: qij})}
        \STATE{$\forall j \in [m]$, update $v_j^*(t)$ based on (\ref{equ:opt_v}).}
        \STATE{$\forall j \in [m]$, if $v_j^*(t)\le \gamma$, construct $R_j=R_c$, else, construct $R_j$ by sample $|R_c|$ contexts in $R_c$. The probability that context $i$ is being sampled is $q_{i,j}/\sum_{i=1}^{|R_c|}q_{i,j}$.}
        \STATE{$\forall j \in [m]$, train model $j$ using the data in $R_j$ for several iterations.}
    \ENDIF
\ENDFOR
\end{algorithmic}\caption{Algorithm for Centroid Approximation Applied to Contextual Bandit.} \label{algo_bootstrap_ci}
\end{algorithm}

\textbf{Training}
For each dataset, we randomly generate 2000 contexts, and for each algorithm, we update the replay memory buffer for each model every 50 contexts, and each model is also updated every 50 contexts. For the standard bootstrap, when updating the replay buffer of each model, we sample 50 i.i.d. contexts with uniform probability from the latest 50 contexts (each model have different realizations of the samples) and add the newly sampled contexts to each model's replay buffer. For the centroid approximation, we update the replay buffer of each model by applying resampling on all the observed contexts up to the current steps. The resampling probability of each context for each model is different and determined by the algorithm. We refer readers to Algorithm \ref{algo_bootstrap_ci} for the detailed procedures. Here we choose freq $=50$ and $M=100$. When at model updating, each model is trained for 100 iterations with batch size 512 using the data from its replay buffer. Following \citet{riquelme2018deep}, we use RMSprop optimizer with learning rate 0.1 for optimizing. When making actions, we sample the prediction head according to $v_k^*(t)$ obtained using the examples between the last two model updates.

Notice that in the implementation, we only need to maintain one common replay buffer and the replay buffer for each model can be implemented by maintaining the number of each context. Thus when sampling batches of context, we simply need to sample the index of the context and refer to the common replay buffer to get the actual data.
}

\subsection{Centroid Approximation for Bootstrap DQN} \label{apx_sec: dqn}

\subsubsection{More Background}
Similar to the bootstrap method in contextual bandit problem, Bootstrap DQN explores using the model uncertainty, which can be assessed via maintaining several models trained with bootstrapped training set. Maintaining several independent models can be very expensive in RL and to reduce the computational cost, Bootstrap DQN uses a multi-head network with a shared base. Each head in the network corresponds to a bootstrap model and the common shared base is thus trained via the union of the bootstrap training set of each head. We train the Bootstrap DQN with standard updating rule for DQN and use Double-DQN \citep{van2016deep} to reduce the overestimate issue.
Notice that our centroid approximation method only changes the memory buffer for each head and thus introduces no conflict to other possible techniques that can be applied to Bootstrap DQN.

\subsubsection{More experiment setup details}
{\color{black}
\textbf{Network Architecture}
Following \citet{osband2016deep}, we considered multi-head network structure with a shared base layer to save the memory. Specifically, we use a fully connected layer with 256 hidden neurons as the shared base and stack two fully connected layers each with 256 hidden neurons as head. Each head in the model can be viewed as one bootstrap particles and in computation, all the bootstrap particles use the same base layer.

\textbf{Training and Evaluation}
For LunarLander-v2, we train the model for 450 episodes with the first 50 episodes used to initialize the common memory buffer. The maximum number of steps within each episode is set to 1000 and we report the moving average reward with window width 100. For Catcher-v0, we train the model for 100 episodes with the first 10 episodes used to initialize the common memory buffer. We set the maximum number of steps within each episodes 2000 and report the moving average reward with window width 25.

For training the Bootstrap DQN, given the current state $x_t$, we sample one particle based on $\{v_j^*\}_{j=1}^m$ and use its policy network to make an action $a_t$ and get the reward $r_t$ and next state $x_{t+1}$. The Q-value of the state action pair $Q(x_t, a_t)$ is estimated by $r_t + \lambda * \hat{Q}(x_{t+1})$, where $\hat{Q}(x_{t+1})$ is the predicted state value by the target network of the sampled particle and $\lambda$ is the discount factor set to be 0.99. At each step, the policy network of all particles are updated using one step gradient descent with Adam optimizer ($\beta=(0.9, 0.999)$ and learning rate 0.001) and mini-batch data (size 64) sampled from its replay buffer. We update target model, each particle's replay buffer and $v_j^*$s every 1000 steps for LunarLander-v2 and every 200 steps for Catcher-v0. The update scheme for replay buffers of each particles and $v_j^*$s is the same as the one in contextual bandit experiment. As the model see significantly larger number of contexts than that in the contextual bandit experiment, to reduce the memory consumption, we set the max capacity of the common replay buffer to 50000 (the oldest data point will be pop out when the size reaches maximum and new data comes in). For training the shared base,
following \citet{osband2016deep}, we adds up all the gradient comes from each head
and normalizes it by the number of heads. Algorithm \ref{algo_bootstrap_dqn} summarizes the whole training pipeline.
}

\begin{algorithm}[t]
\begin{algorithmic}[1]
\STATE{Obtain a randomly initialized $\th^*_j(0)$, $j\in[m]$. (For the $j$-th particle, both of its target and policy network use the same initialization.)}
\STATE{Initialize a common replay buffer $R_c = \emptyset$ recording all the observed contexts.}
\STATE{For each head, initialize its own replay buffer $R_j = \emptyset$ that is used for training.}
\FOR{$t \in \text{number of total episodes}$}
    \WHILE {not at terminal state and the number of steps does not exceed the threshold}
        \STATE{Obtain the $t$-th context $x_t$.}
        \STATE{Sample an head based on $\{v_j^*\}$ to make action $a_t$ and get $Q(x_t, a_t)$ using the reward $r_t$ and the prediction of the corresponding target network.}
        \STATE{Update the common replay buffer by $R_{c}\leftarrow R_{c}\cup\{(x_t, Q(x_t, a_t))\}$}
        \STATE{$\forall j \in [m]$, update its policy network by one step gradient descent using the data from its reply buffer.}
        \STATE{// Update $\w_h$ and $R_j$ and target network every a few iterations.}
            \IF{$t$ \text{mod freq} $== 0$}
                \STATE{$\forall j\in[m]$, calculate $\LL(\th^*_j(t))$ defined in (\ref{equ: datawise loss}) for all the contexts in $R_c$.}
                \STATE{Generate $M$ sets of random weights $\{\w_h\}_{h=1}^M$ of contexts in $R_c$ from $\pi$.}
                \STATE{$\forall h \in [M]$ and $j \in [m]$, calculate $\LL_{\w_h}(\th^*_j(t))=\w_h^T\LL(\th_j^*(t))$.}
                \STATE{$\forall h \in [M]$, calculate $u_{\w_h}(t)$ defined in (\ref{equ: uwh}) for each $h$.}
                \STATE{$\forall i \in [|R_c|]$ and $j \in [m]$, calculate $q_{i,j}$ by (\ref{equ: qij})}
                \STATE{$\forall j \in [m]$, update $v_j^*(t)$ based on (\ref{equ:opt_v}).}
                \STATE{$\forall j \in [m]$, if $v_j^*(t)\le \gamma$, construct $R_j=R_c$, else, construct $R_j$ by sample $|R_c|$ contexts in $R_c$. The probability that context $i$ is being sampled is $q_{i,j}/\sum_{i=1}^{|R_c|}q_{i,j}$.}
                \STATE{$\forall j \in [m]$, update the $j$-th target network by loading the weights of the $j$-th policy network.}
            \ENDIF
    \ENDWHILE
\ENDFOR
\end{algorithmic}\caption{Algorithm for Centroid Approximation Applied to DQN.} \label{algo_bootstrap_dqn}
\end{algorithm}

\subsection{Bootstrap Ensemble Model} \label{apx_sec: bagging}

\begin{figure}[t]
\begin{centering}
\includegraphics[scale=0.3]{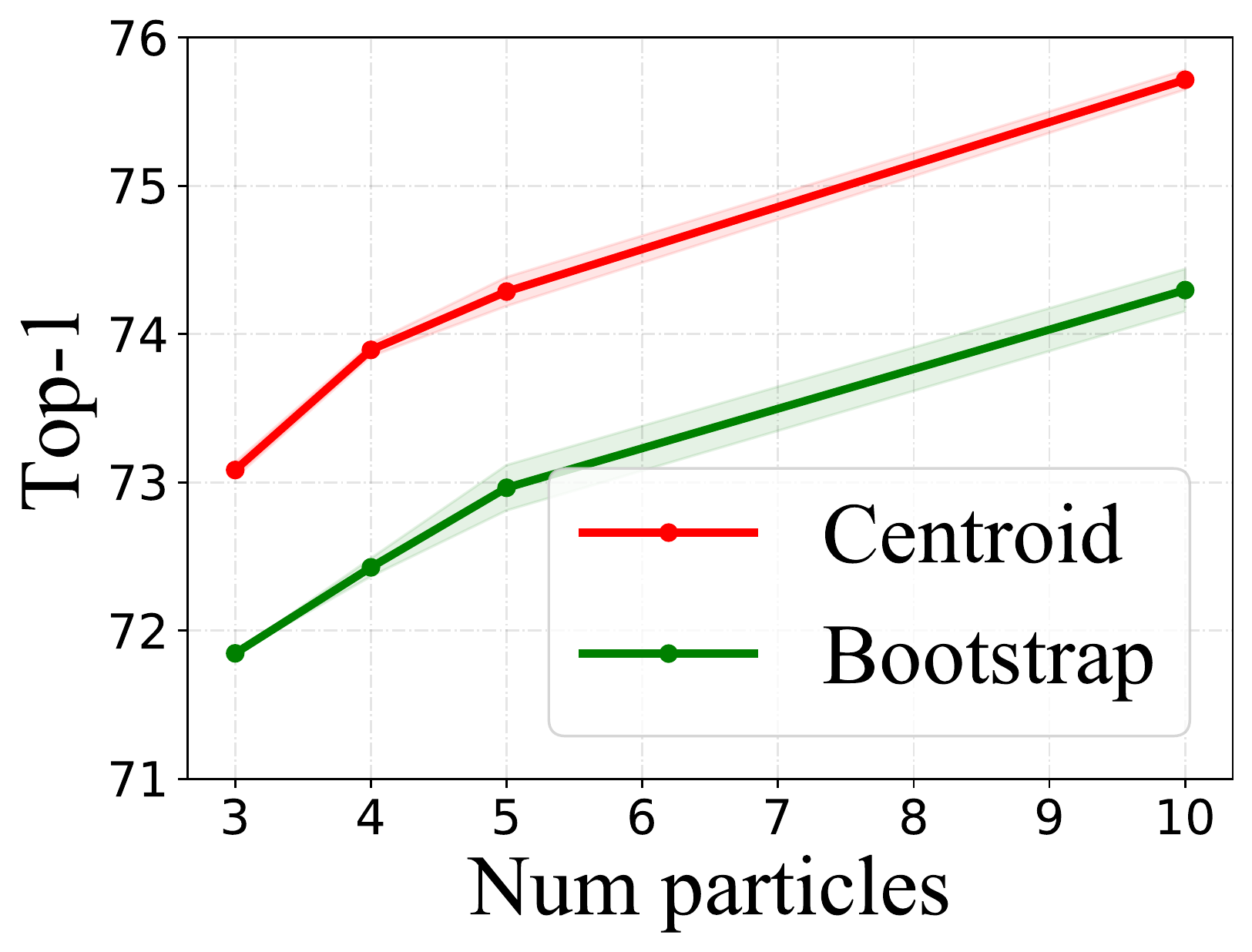}\includegraphics[scale=0.3]{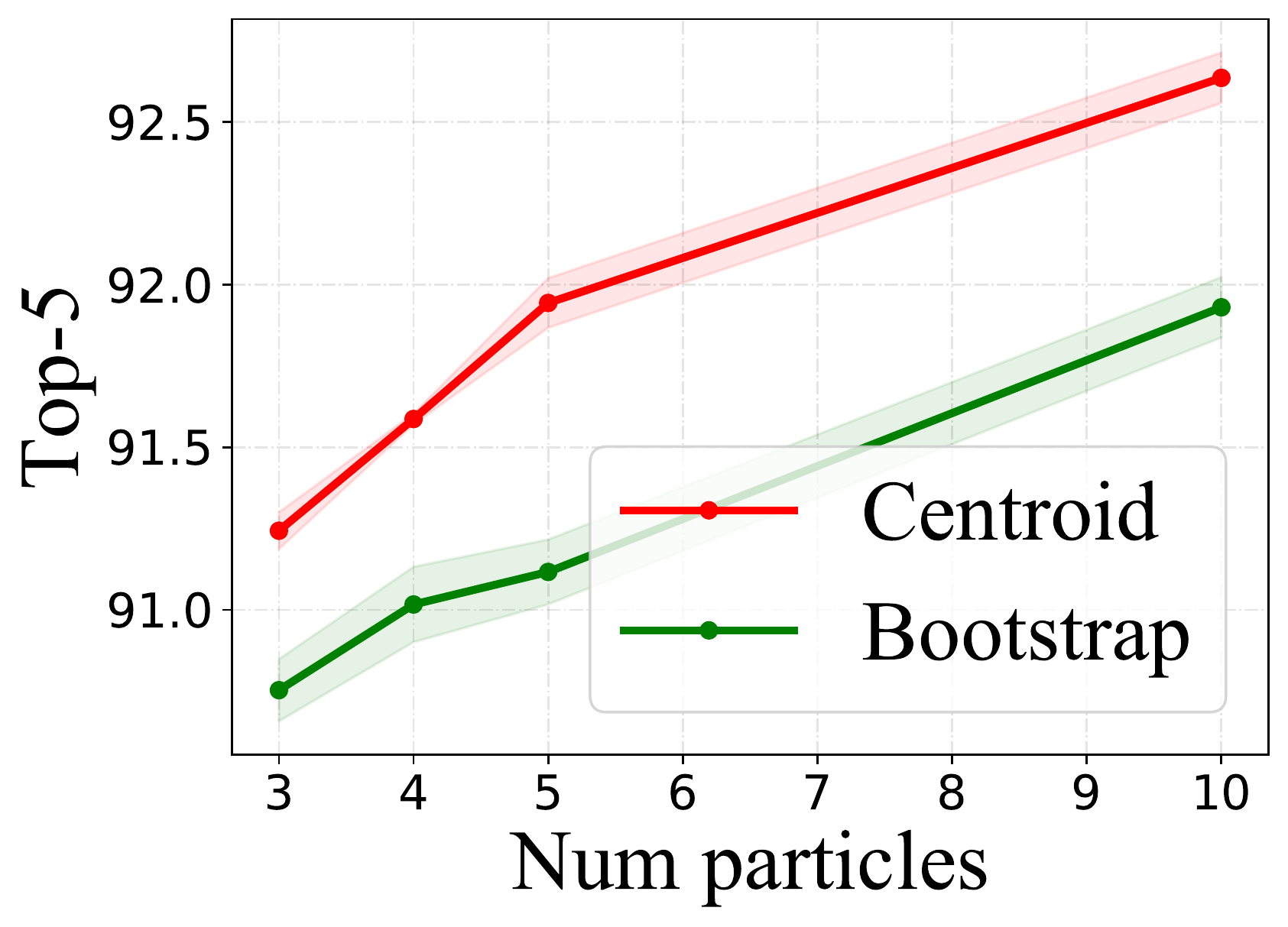} 
\caption{Results on ensemble modeling with bootstrap using Vgg16 on CIFAR-100.} \label{fig: ensemble}
\par\end{centering}
\end{figure}

Ensemble of deep neural networks have been successfully used to boost predictive performance \citep{lakshminarayanan2016simple}. In this experiment, we consider using an ensemble of deep neural network trained on different bootstrapped training set, which is also known as a popular strategy called \emph{bagging}.

We consider image classification task on CIFAR-100 and use standard VGG-16 \citep{simonyan2014very} with batch normalization. We apply a standard training pipeline. We train the bootstrap model for 160 epochs using SGD optimizer with 0.9 momentum and batchsize 128. The learning rate is initialized to be 0.1 and is decayed by a factor of 10 at epoch 80 and 120. We start to apply the centroid approximation at epoch 120 (thus the centroid is initialized with 120 epochs' training). We generate the bootstrap training set for each centroid every epoch using the proposed centroid approximation method. We consider $m=3,4,5,10$ ensembles and use $\gamma = 0.5/m$. We repeat the experiment for 3 random trials and report the averaged top1 and top5 accuracy with the standard deviation of the mean estimator. Algorithm

Figure \ref{fig: ensemble} summarizes the result. Overall, increasing $m$ is able to improve the predictive performance and with the same number of models, our centroid approximation consistently improves over standard bootstrap ensembles.

\subsection{Ablation Study} \label{apx_sec: ablation}
We study the effectiveness of using (\ref{upadte: centroid}) to modify the gradient of centroid with $v_k^*(t)\le\gamma$. We consider the setting $\gamma = 0$ (no modification) and $\gamma = m$ (always modify, equivalent to no bootstrap uncertainty) and applied the method on the mushroom dataset in the contextual bandit problem. Table \ref{table: ablation} shows that (i) modifying the gradient of centroid with small $v_k^*(t)$ using do improve the overall result; (2) bootstrap uncertainty is important for exploration.

\begin{table}[t]
\centering{}%
\begin{tabular}{c|ccc}
\toprule 
 \small{\#Particle} & $\gamma=0$ & $\gamma=0.5$ & $\gamma=m$\tabularnewline
\hline 
3 & $3480.0\pm120$ & $3702.7\pm89.8$ & $3467.7\pm115$\tabularnewline
4 & $3461.9\pm126$ & $3723.1\pm78.7$ & $3600.0\pm69.3$\tabularnewline
5 & $3586.5\pm64.5$ & $3799.6\pm84.2$ & $3647.3\pm64.5$\tabularnewline
10 & $3785.0\pm59.1$ & $3796.9\pm36.1$ & $3742.7\pm86.8$\tabularnewline
\bottomrule 
\end{tabular}\caption{Ablation study on $\gamma$.} \label{table: ablation}
\vspace{-0.5cm}
\end{table}

\subsection{Computation overhead} \label{apx_sec: overhead}
Our main goal is not to decrease the training cost but improve the quality of bootstrap partical distribution when $m$ is small. On the other hand, as discussed in Section \ref{sec: method}, our method actually only introduces a little computation overhead while much improves the quality of the particles, which is another advantage of our method. To demonstrate this, we summarize the training time of our centroid approximation and naive bootstrap in Table \ref{tbl: time} and \ref{tbl: time2}. Results are based on the average of 3 runs. Note that in the Bootstrap DQN application, as the number of iterations in each episode depends on the executed action by the algorithm, our centroid approximation can have smaller training time in Catcher experiment. In summary, our method only introduces about $7\%$ computational overhead even with an naive implementation. 

\begin{table}
\centering{}%
\begin{tabular}{c|c|cccc}
\toprule 
\multicolumn{2}{c|}{Wall clock time} & \multirow{2}{*}{$m=3$} & \multirow{2}{*}{$m=4$} & \multirow{2}{*}{$m=5$} & \multirow{2}{*}{$m=10$}\tabularnewline
\multicolumn{2}{c|}{Bootstrap/Centroid} &  &  &  & \tabularnewline
\hline 
\multirow{3}{*}{Contextual Bandit} & Mushroom & $33.0/34.9$ & $40.7/42.6$ & $47.4/51.8$ & $101/108$\tabularnewline
 & Statlog & $22.1/22.9$ & $30.2/31.1$ & $38.9/40.0$ & $74.3/75.7$\tabularnewline
 & Financial & $23.1/24.8$ & $30.6/32.1$ & $39.5/40.7$ & $75.1/78.9$\tabularnewline
\hline 
Ensemble & CIFAR-100 & $187/199$ & $245/262$ & $303/332$ & $571/563$\tabularnewline
\bottomrule 
\end{tabular}\caption{Training time comparison for contextual banidt and ensemble learning application. The time unit is second for Contextual Bandit and minute for Ensemble model.} \label{tbl: time}
\end{table}

\begin{table}
\centering{}%
\begin{tabular}{c|c|cc}
\toprule
\multicolumn{2}{c|}{Wall clock time} & \multirow{2}{*}{$m=2$} & \multirow{2}{*}{$m=5$}\tabularnewline
\multicolumn{2}{c|}{Bootstrap/Centroid} &  & \tabularnewline
\hline 
\multirow{2}{*}{Bootstrap DQN} & LunarLand & $40.5/41.8$ & $92.8/107$\tabularnewline
 & Catcher & $141/121$ & $286/264$\tabularnewline
\bottomrule 
\end{tabular}\caption{Training time comparison for bootstrap dqn. The time unit is minuts.} \label{tbl: time2}
\end{table}

\section{Proof}
We also show Theorem \ref{thm: taylor}, which gives a formal characterization of the Taylor approximation intuition introduced in (\ref{equ: taylor_intuition}). 

\begin{theorem} \label{thm: taylor}
Under Assumption 1 and 2, when $n$ is sufficiently large, we have 
\begin{eqnarray*}
&\!\!\!\!\L_{\w}(\th)=\L_{\w}(\hat{\th}_{\w})+\frac{1}{2} 
\left(\th-\hat{\th}_{\w}\right)^{\top}\nabla_{\th}^{2}\L_{\infty}(\th_{0})\left(\th-\hat{\th}_{\w}\right)
\\
&+O_{p}\left(||\th-\hat{\th}_{\w}||^{2}(n^{-1/2}+||\th-\hat{\th}_{\w}||)\right).
\end{eqnarray*}
Here the stochastic boundedness is taken w.r.t. the training data and $\w$.
\end{theorem}

In the proof, we may use $c$ to represent some absolute constant, which may vary in different lines.
\subsection{Proof of Theorem \ref{thm: taylor}}

With the fact that $\nabla_{\w}\L(\hat{\th}_{\w})=0$ and under assumption \ref{assumption: bound}, using Taylor expansion, we have
\[
\L_{\w}(\th)=\L_{\w}(\hat{\th}_{\w})+\frac{1}{2}\left(\th-\hat{\th}_{\w}\right)^{\top}\nabla_{\th}^{2}\L_{\w}(\hat{\th}_{\w})\left(\th-\hat{\th}_{\w}\right)+O\left(\left\Vert \th-\hat{\th}_{\w}\right\Vert ^{3}\right).
\]
Notice that 
\begin{align*}
 & \left|\left(\th-\hat{\th}_{\w}\right)^{\top}\left(\nabla_{\th}^{2}\L_{\w}(\hat{\th}_{\w})-\nabla_{\th}^{2}\L_{\infty}(\th_{0})\right)\left(\th-\hat{\th}_{\w}\right)\right|\\
\le & \left\Vert \th-\hat{\th}_{\w}\right\Vert ^{2}\left\Vert \nabla_{\th}^{2}\L_{\w}(\hat{\th}_{\w})-\nabla_{\th}^{2}\L_{\infty}(\th_{0})\right\Vert \\
\le & \left\Vert \th-\hat{\th}_{\w}\right\Vert ^{2}\left(\left\Vert \nabla_{\th}^{2}\L_{\w}(\hat{\th}_{\w})-\nabla_{\th}^{2}\L_{\w}(\th_{0})\right\Vert +\left\Vert \nabla_{\th}^{2}\L_{\w}(\th_{0})-\nabla_{\th}^{2}\L_{\infty}(\th_{0})\right\Vert \right)\\
\le & \left\Vert \th-\hat{\th}_{\w}\right\Vert ^{2}\left(C\left\Vert \hat{\th}_{\w}-\th_{0}\right\Vert +\left\Vert \nabla_{\th}^{2}\L_{\w}(\th_{0})-\nabla_{\th}^{2}\L_{\infty}(\th_{0})\right\Vert _{F}\right),
\end{align*}
where we denote the Frobenius norm as $\left\Vert \cdot\right\Vert _{F}$.
With assumption \ref{assumption: asy-normal}, we have $\left\Vert \hat{\th}_{\w}-\th_{0}\right\Vert =O_{p}(n^{-1/2})$.
By applying centroid limit theorem and delta method to $\left|\nabla_{\th_{ij}}^{2}\L_{\w}(\th_{0})-\nabla_{\th_{ij}}^{2}\L_{\infty}(\th_{0})\right|$
for every pair $i,j\in[d]$, we have $\left\Vert \nabla_{\th}^{2}\L_{\w}(\th_{0})-\nabla_{\th}^{2}\L_{\infty}(\th_{0})\right\Vert _{F}=O_{p}(n^{-1/2})$.
Thus we obtained the desired result.

\subsection{Proof of Theorem \ref{thm: stable_main}}

Given any radius $r$ and $\epsilon>0$, with sufficiently large $n$, we have 
\[
\P\left(\left\Vert \hat{\th}_{w}-\th_{0}\right\Vert \ge r\right)\le\exp(-\lambda nr^{2}) {\color{black}+ \epsilon/m},
\]
for $\lambda=\frac{1}{4}\lambda_{\max}(A)^{-1}$. Here the probability
is the jointly probability of bootstrap weight and training data.
Thus, given any $r$, under the assumption that $\th_{j}^{*}(0)$
is initialized via sampling $\hat{\th}_{w}$, then we have 
\[
\P\left(\cup_{j\in[m]}\left\{ \left\Vert \th_{j}^{*}(0)-\th_{0}\right\Vert \ge r\right\} \right)\le\sum_{j\in[m]}\P\left(\left\Vert \th_{j}^{*}(0)-\th_{0}\right\Vert \ge r\right)\le m\exp(-\lambda nr^{2}) + {\color{black}\epsilon}.
\]
We proof by induction. Given any $\{\th_{j}\}_{j=1}^{m}$, define
\[
R_{k,r}=\mathbb{I}\left\{ w\in\text{supp(\ensuremath{\pi})}:\arg\min_{j\in[m]}\L_{\w}(\th_{j})=k\ \text{and}\ \left\Vert \hat{\th}_{w}-\th_{0}\right\Vert \le r\right\} .
\]
Suppose at iteration $t$, we have $\left\Vert \th_{k}^{*}(t)-\th_{0}\right\Vert \le\frac{c\alpha\sqrt{\frac{\log n}{n}}}{\lambda_{0}\gamma}$
for some constant $c$ and $\lambda_{0}$, which we denote as the minimum eigenvalue
of $\nabla_{\th}^{2}\L_{\infty}(f_{\th_{0}})$. Now at iteration $t$,
we have two cases.

\paragraph{Case 1: $\protect\E_{\pi}R_{k,\infty}\ge\gamma$}

Suppose that at iteration $t$, for $k$ such that $\E_{\pi}R_{k,\infty}\ge\gamma$,
and $\left\Vert \th_{k}^{*}(t)-\th_{0}\right\Vert =q_{k}$, we have
the following property:
\begin{align*}
\left\Vert \th_{k}^{*}(t+1)-\th_{0}\right\Vert ^{2} & =\left\Vert \th_{k}^{*}(t)-\frac{\epsilon_{t}}{\E_{\pi}R_{k,\infty}}\E_{\pi}\left[\nabla_{\th}\L_{\w}(\th_{k}^{*}(t))R_{k,\infty}\right]-\th_{0}\right\Vert ^{2}\\
 & =\left\Vert \th_{k}^{*}(t)-\th_{0}\right\Vert ^{2}-\frac{2\epsilon_{t}}{\E_{\pi}R_{k,\infty}}\left\langle \th_{k}^{*}(t)-\th_{0},\E_{\pi}\left[\nabla_{\th}\L_{\w}(\th_{k}^{*}(t))R_{k,\infty}\right]\right\rangle +\epsilon_{t}^{2}\left\Vert \E_{\pi}\left[\nabla_{\th}\L_{\w}(\th_{k}^{*}(t))R_{k,\infty}\right]\right\Vert ^{2}.
\end{align*}
Notice that 
\begin{align*}
\E_{\pi}\left[\nabla_{\th}\L_{\w}(\th_{k}^{*}(t))R_{k,q_{k}}\right] & \overset{(1)}{=}\E_{\pi}\left[\nabla_{\th}^{2}\L_{\w}(\hat{\th}_{w})(\th_{k}^{*}(t)-\hat{\th}_{w})R_{k,q_{k}}\right]+o\left(q_{k}^{2}\right)\\
 & \overset{(2)}{=}\E_{\pi}\left[\nabla_{\th}^{2}\L_{\w}(\hat{\th}_{w})(\th_{k}^{*}(t)-\th_{0})R_{k,q_{k}}\right]+O\left(q_{k}^{2}\right)\\
 & \overset{(3)}{=}\E_{\pi}\left[\nabla_{\th}^{2}\L_{\w}(\th_{0})(\th_{k}^{*}(t)-\th_{0})R_{k,q_{k}}\right]+O\left(q_{k}^{2}\right)
\end{align*}
Here $(1)$ is obtained via applying Taylor expansion on $\nabla_{\th}\L_{\w}(\th_{k}^{*}(t))$
at $\hat{\th}_{\w}$. $(2)$ is by assumption \ref{assumption: bound} and \ref{assumption: asy-normal}. $(3)$ is
by assumption \ref{assumption: bound}. We thus
have
\begin{align*}
& -\left\langle \th_{k}^{*}(t)-\th_{0},\E_{\pi}\left[\nabla_{\th}\L_{\w}(\th_{k}^{*}(t))R_{k,\infty}\right]\right\rangle  
\\
\le & -\left\langle \th_{k}^{*}(t)-\th_{0},\E_{\pi}\left[\nabla_{\th}\L_{\w}(\th_{k}^{*}(t))R_{k,q_{k}}\right]\right\rangle +\left\Vert \th_{k}^{*}(t)-\th_{0}\right\Vert \left\Vert \E_{\pi}\nabla_{\th}\L_{\w}(\th_{k}^{*}(t))(1-R_{k,q_{k}})\right\Vert \\
\le & -\left\langle \th_{k}^{*}(t)-\th_{0},\E_{\pi}\left[\nabla_{\th}\L_{\w}(\th_{k}^{*}(t))R_{k,q_{k}}\right]\right\rangle +cq_{k}\exp(-\lambda nq_{k}^{2})\\
\le & -\E_{\pi}R_{k,q_{k}}(\th_{k}^{*}(t)-\th_{0})^{\top}\nabla_{\th}^{2}\L_{\w}(\th_{0})(\th_{k}^{*}(t)-\th_{0})+cq_{k}\exp(-\lambda nq_{k}^{2})+O\left(q_{k}^{3}\right).
\end{align*}
Notice that with sufficiently large $n$, with central limit theorem,
we have
\begin{align*}
 & -\E_{\pi}R_{k,q_{k}}(\th_{k}^{*}(t)-\th_{0})^{\top}\nabla_{\th}^{2}\L_{\w}(\th_{0})(\th_{k}^{*}(t)-\th_{0})\\
\le & \left\Vert \th_{k}^{*}(t)-\th_{0}\right\Vert ^{2}\E_{\pi}\left\Vert \nabla_{\th}^{2}\L_{\w}(\th_{0})-\nabla_{\th}^{2}\L_{\w}(\th_{0})\right\Vert -\E_{\pi}R_{k,q_{k}}(\th_{k}^{*}(t)-\th_{0})^{\top}\nabla_{\th}^{2}\L_{\infty}(\th_{0})(\th_{k}^{*}(t)-\th_{0})\\
= & -\E_{\pi}R_{k,q_{k}}(\th_{k}^{*}(t)-\th_{0})^{\top}\nabla_{\th}^{2}\L_{\infty}(\th_{0})(\th_{k}^{*}(t)-\th_{0})+O(n^{-1/2}).
\end{align*}
This gives that 
\begin{align*}
 & -\left\langle \th_{k}^{*}(t)-\th_{0},\E_{\pi}\left[\nabla_{\th}\L_{\w}(\th_{k}^{*}(t))R_{k,\infty}\right]\right\rangle \\
\le & -\E_{\pi}R_{k,q_{k}}(\th_{k}^{*}(t)-\th_{0})^{\top}\nabla_{\th}^{2}\L_{\infty}(\th_{0})(\th_{k}^{*}(t)-\th_{0})+cq_{k}\exp(-\lambda nq_{k}^{2})+O\left(q_{k}^{3}+q_{k}n^{-1/2}\right)\\
\le & -\lambda_{0}\E_{\pi}R_{k,q_{k}}\left\Vert \th_{k}^{*}(t)-\th_{0}\right\Vert ^{2}+cq_{k}\exp(-\lambda nq_{k}^{2})+O\left(q_{k}^{3}+q_{k}n^{-1/2}\right).
\end{align*}
Use the above estimation, we have 
\begin{align*}
& \left\Vert \th_{k}^{*}(t+1)-\th_{0}\right\Vert ^{2} 
\\
\le & \left\Vert \th_{k}^{*}(t)-\th_{0}\right\Vert ^{2}-2\epsilon_{t}\lambda_{\min}\frac{\E_{\pi}R_{k,q_{k}}}{\E_{\pi}R_{k,\infty}}\left\Vert \th_{k}^{*}(t)-\th_{0}\right\Vert ^{2}\\
+ & 2c\epsilon_{t}q_{k}\exp(-\lambda nq_{k}^{2})/\E_{\pi}R_{k,\infty}+O\left(\epsilon_{t}(q_{k}^{3}+q_{k}n^{-1/2})/\E_{\pi}R_{k,\infty}+\epsilon_{t}^{2}\right)\\
\le & \left\Vert \th_{k}^{*}(t)-\th_{0}\right\Vert ^{2}+\text{\ensuremath{\frac{\epsilon_{t}}{\E_{\pi}R_{k,\infty}}\left(-2\lambda_{0}\E_{\pi}R_{k,q_{k}}\left\Vert \th_{k}^{*}(t)-\th_{0}\right\Vert ^{2}-2cq_{k}\exp(-\lambda nq_{k}^{2})+O\left(q_{k}^{3}+\epsilon_{t}+q_{k}n^{-1/2}\right)\right).}}
\end{align*}
Notice that by choosing $\alpha>\sqrt{1/(2\lambda)}$ and $\epsilon_{t}=O(n^{-1})$,
with sufficiently large $n$, when 
\[
\left\Vert \th_{k}^{*}(t)-\th_{0}\right\Vert \ge\frac{c\alpha\sqrt{\frac{\log n}{n}}}{\lambda_{0}\E_{\pi}R_{k,q_{k}}}
\]
for some constant $c$, we have 
\[
\left\Vert \th_{k}^{*}(t+1)-\th_{0}\right\Vert \le\left\Vert \th_{k}^{*}(t)-\th_{0}\right\Vert .
\]
Thus $\left\Vert \th_{k}^{*}(t+1)-\th_{0}\right\Vert \le\frac{c\alpha\sqrt{\frac{\log n}{n}}}{\lambda_{0}\E_{\pi}R_{k,\infty}}\le\frac{c\alpha\sqrt{\frac{\log n}{n}}}{\lambda_{0}\gamma}$
for some constant $c$.

\paragraph{Case 2: $\protect\E_{\pi}R_{k,\infty}\le\gamma$}

In this case, we have

\begin{align*}
\left\Vert \th_{k}^{*}(t+1)-\th_{0}\right\Vert ^{2} & =\left\Vert \th_{k}^{*}(t)-\epsilon_{t}\nabla_{\th}\L(f_{\th_{k}^{*}(t)})-\th_{0}\right\Vert ^{2}\\
 & =\left\Vert \th_{k}^{*}(t)-\th_{0}\right\Vert ^{2}-2\epsilon_{t}\left\langle \th_{k}^{*}(t)-\th_{0},\nabla_{\th}\L(\th_{k}^{*}(t))\right\rangle +\epsilon_{t}^{2}\left\Vert \nabla_{\th}\L(f_{\th_{k}^{*}(t)})\right\Vert ^{2}.
\end{align*}
Notice that 
\begin{align*}
-\left\langle \th_{k}^{*}(t)-\th_{0},\nabla_{\th}\L(\th_{k}^{*}(t))\right\rangle  & \le-\left\langle \th_{k}^{*}(t)-\th_{0},\nabla_{\th}^{2}\L(\th_{0})\left(\th_{k}^{*}(t)-\th_{0}\right)\right\rangle -\left\langle \th_{k}^{*}(t)-\th_{0},\nabla_{\th}\L(f_{\th_{0}})\right\rangle +o(||\th_{k}^{*}(t)-\th_{0}||^{3})\\
 & =-\left(\th_{k}^{*}(t)-\th_{0}\right)^{\top}\nabla_{\th}^{2}\L_{\infty}(\th_{0})\left(\th_{k}^{*}(t)-\th_{0}\right)+o(||\th_{k}^{*}(t)-\th_{0}||^{3})+O_{p}(n^{-1/2})||\th_{k}^{*}(t)-\th_{0}||.
\end{align*}
This gives that 
\[
\left\Vert \th_{k}^{*}(t+1)-\th_{0}\right\Vert ^{2}\le\left\Vert \th_{k}^{*}(t)-\th_{0}\right\Vert ^{2}-2\epsilon_{t}\lambda_{0}\left\Vert \th_{k}^{*}(t)-\th_{0}\right\Vert ^{2}+o(\epsilon_{t}||\th_{k}^{*}(t)-\th_{0}||^{3}+\epsilon_{t}^{2})+O_{p}(n^{-1/2})\epsilon_{t}||\th_{k}^{*}(t)-\th_{0}||.
\]
With $\epsilon_{t}=O(n^{-1})$ and sufficiently large $n$, when 
\[
\left\Vert \th_{k}^{*}(t)-\th_{0}\right\Vert \ge\frac{c\alpha\sqrt{\frac{\log n}{n}}}{\lambda_{0}\gamma},
\]
we have $\left\Vert \th_{k}^{*}(t+1)-\th_{0}\right\Vert ^{2}\le\left\Vert \th_{k}^{*}(t)-\th_{0}\right\Vert ^{2}$.

By these two cases, we conclude that $\left\Vert \th_{k}^{*}(t+1)-\th_{0}\right\Vert \le\frac{c\alpha\sqrt{\frac{\log n}{n}}}{\lambda_{0}\gamma}$
for any $t$, when $\left\Vert \th_{k}^{*}(0)-\th_{0}\right\Vert \le\frac{c\alpha\sqrt{\frac{\log n}{n}}}{\lambda_{0}\gamma}$.
We thus conclude that, for any $\alpha>\sqrt{1/(2\lambda)}$ and $\epsilon>0$, when $n$ is sufficiently large, with
probability at least $1-m\exp\left(-\lambda\frac{c\alpha^{2}\log n}{\lambda_{0}^{2}\gamma^{2}}\right) - \epsilon$,
we have 
\[
\max_{j\in[m]}\sup_{t}\left\Vert \th_{j}^{*}(t)-\th_{0}\right\Vert \le\frac{c\alpha\sqrt{\frac{\log n}{n}}}{\lambda_{0}\gamma}.
\]

\subsection{Proof for Theorem \ref{thm: good_surrogate}}
Notice that 
\begin{align*}
\L_{\w}(\th_{j}^{*}(t))-\L_{\w}(\hat{\th}_{\w}) & =\frac{1}{2}\left(\th_{j}^{*}(t)-\hat{\th}_{\w}\right)^{\top}\nabla_{\th}^{2}\L_{\w}(\hat{\th}_{\w})\left(\th_{j}^{*}(t)-\hat{\th}_{\w}\right)+O(\left\Vert \th_{j}^{*}(t)-\hat{\th}_{\w}\right\Vert ^{3})\\
 & =\frac{1}{2}\left(\th_{j}^{*}(t)-\hat{\th}_{\w}\right)^{\top}\nabla_{\th}^{2}\L_{\w}(\th_{0})\left(\th_{j}^{*}(t)-\hat{\th}_{\w}\right)\\
 & + O(\left\Vert \th_{j}^{*}(t)-\hat{\th}_{\w}\right\Vert ^{3})+O(\left\Vert \th_{j}^{*}(t)-\hat{\th}_{\w}\right\Vert ^{2}\left\Vert \hat{\th}_{\w}-\th_{0}\right\Vert )\\
 & =\frac{1}{2}\left\Vert \th_{j}^{*}(t)-\hat{\th}_{\w}\right\Vert _{D}^{2}+\left\Vert \nabla_{\th}^{2}\L_{\w}(\th_{0})-\nabla_{\th}^{2}\L_{\infty}(\th_{0})\right\Vert \left\Vert \th_{j}^{*}(t)-\hat{\th}_{\w}\right\Vert ^{2}+ O(\left\Vert \th_{j}^{*}(t)-\hat{\th}_{\w}\right\Vert ^{3})\\
 & + O(\left\Vert \th_{j}^{*}(t)-\hat{\th}_{\w}\right\Vert ^{2}\left\Vert \hat{\th}_{\w}-\th_{0}\right\Vert )
\end{align*}

Given $\w$, we define $u_{\w}=\arg\min_{j\in[m]}\left\Vert \th_{j}^{*}(t)-\hat{\th}_{w}\right\Vert _{D}^{2}$.
For any $\alpha>\sqrt{1/(2\lambda)}$ and $\epsilon>0$, when $n$ is sufficiently large, with probability at least $1-m\exp\left(-\lambda\frac{c\alpha^{2}\log n}{\lambda_{0}^{2}\gamma^{2}}\right)-\epsilon$,
we have 
\begin{align*}
 & \frac{1}{2}\E_{\w\sim\pi}\left[\min_{j\in[m]}\left\Vert \th_{j}^{*}(t)-\hat{\th}_{w}\right\Vert _{D}^{2}\right]\\
= & \frac{1}{2}\E_{\w\sim\pi}\left[\left\Vert \th_{u_{\w}}^{*}-\hat{\th}_{w}\right\Vert _{D}^{2}\right]\\
\ge & \E_{\w\sim\pi}\left[\L_{\w}(\th_{u_{\w}}^{*})-\L_{\w}(\hat{\th}_{\w})-c\left\Vert \th_{u_{\w}}^{*}-\hat{\th}_{\w}\right\Vert ^{2}\left(\left\Vert \hat{\th}_{\w}-\th_{0}\right\Vert +\left\Vert \nabla_{\th}^{2}\L_{\w}(\th_{0})-\nabla_{\th}^{2}\L_{\infty}(\th_{0})\right\Vert
+\left\Vert \th_{u_{\w}}^{*}-\hat{\th}_{\w}\right\Vert
\right)\right]\\
\ge & \E_{\w\sim\pi}\left[\L_{\w}(\th_{u_{\w}}^{*})\right]-\E_{\w\sim\pi}\left[\L_{\w}(\hat{\th}_{\w})\right]-c\left(\frac{\alpha\sqrt{\log n}}{\lambda_{0}\gamma n^{3/2}}\right)\\
= & \E_{\w\sim\pi}\left[\min_{j\in[m]}\L_{\w}(\th_{j}^{*}(t))\right]-\E_{\w\sim\pi}\left[\L_{\w}(\hat{\th}_{\w})\right]-c\left(\frac{\alpha\sqrt{\log n}}{\lambda_{0}\gamma n^{3/2}}\right).
\end{align*}
Similarly, we also have, with probability at least $1 - m\exp\left(-\lambda\frac{c\alpha^{2}\log n}{\lambda_{0}^{2}\gamma^{2}}\right)$,
\[
\E_{\w\sim\pi}\left[\min_{j\in[m]}\L_{\w}(\th_{j}^{*}(t))\right]-\E_{\w\sim\pi}\left[\L_{\w}(\hat{\th}_{\w})\right]\ge\frac{1}{2}\E_{\w\sim\pi}\left[\min_{j\in[m]}\left\Vert \th_{j}^{*}(t)-\hat{\th}_{w}\right\Vert _{D}^{2}\right]-c\left(\frac{\alpha\sqrt{\log n}}{\lambda_{0}\gamma n^{3/2}}\right).
\]
Notice that the above bound holds uniformly for all $j\in[m]$ and
any iteration $t$, which implies that with probability at least $1-2m\exp\left(-\lambda\frac{c\alpha^{2}\log n}{\lambda_{0}^{2}\gamma^{2}}\right)-2\epsilon$,
we have 
\[
\sup_{t\ge0}\left|\E_{\w\sim\pi}[\min_{j\in[m]}\L_{\w}(\th_{j}^{*}(t))]-B-\E_{\w\sim\pi}[\min_{j\in[m]}||\th_{j}^{*}(t)-\hat{\th}_{\w}||_{D}^{2}]/2\right| \le c\left(\frac{\alpha\sqrt{\log n}}{\lambda_{0}\gamma n^{3/2}}\right).
\]


\end{document}